\documentclass[lettersize,journal]{IEEEtran}
\usepackage{amsmath,amsfonts}
\usepackage{array}
\usepackage[caption=false,font=normalsize,labelfont=sf,textfont=sf]{subfig}
\usepackage{textcomp}
\usepackage{stfloats}
\usepackage{url}
\usepackage{verbatim}
\usepackage{graphicx}
\def\BibTeX{{\rm B\kern-.05em{\sc i\kern-.025em b}\kern-.08em
    T\kern-.1667em\lower.7ex\hbox{E}\kern-.125emX}}
\usepackage{balance}

\usepackage{booktabs}
\usepackage{multirow}
\usepackage{colortbl}
\usepackage{graphicx}
\usepackage{subcaption}
\usepackage{listings}
\usepackage{algorithmicx}
\usepackage{minted}
\usepackage{pythonhighlight}
\usepackage{footmisc} 
\usepackage[colorlinks,linkcolor=red]{hyperref}

\newcommand{\re}[1]{\textcolor{black}{#1}}

\begin{document}

\title{AdalReID: Auto-selected Knowledge Adapters for Lifelong Person Re-identification}


\author{
Xuelin Qian,~\IEEEmembership{Member,~IEEE}, Ruiqi Wu, 
Zhongling Huang,~\IEEEmembership{Member,~IEEE}, Dingwen Zhang,~\IEEEmembership{Member,~IEEE} and Junwei Han,~\IEEEmembership{Fellow,~IEEE} 
}

\markboth{Journal of \LaTeX\ Class Files,~Vol.~18, No.~9, September~2020}%
{AdalReID: Auto-selected Knowledge Adapters for Lifelong Person Re-identification}

\maketitle

\begin{abstract}
Lifelong Person Re-Identification (LReID) extends traditional ReID by requiring systems to continually learn from non-overlapping datasets across different times and locations, adapting to new identities while preserving knowledge of previous ones.
Existing approaches, either rehearsal-free or rehearsal-based, still suffer from the problem of catastrophic forgetting since they try to cram diverse knowledge into one fixed model. To overcome this limitation, we introduce a novel framework AdalReID, that adopts knowledge adapters
and a parameter-free auto-selection mechanism for lifelong learning. Concretely, we incrementally build distinct adapters to learn domain-specific knowledge at each step, which can effectively learn and preserve knowledge across different datasets. 
Meanwhile, the proposed auto-selection strategy adaptively calculates the knowledge similarity between the input set and the adapters. On the one hand, the appropriate adapters are selected for the inputs to process ReID, and on the other hand, the knowledge interaction and fusion between adapters are enhanced to improve the generalization ability of the model.
Extensive experiments are conducted to demonstrate the superiority of our AdalReID, which significantly beats SOTAs by about 5\% and 3\% mAP on seen and unseen domains.
\end{abstract}

\begin{IEEEkeywords}
Lifelong person re-identification, knowledge adapters, parameter-free auto-selection
\end{IEEEkeywords}

\section{Introduction \label{sec:intro}}
\IEEEPARstart{P}{erson} re-identification (ReID) aims to identify individuals across different camera views. 
While existing ReID methods~\cite{qian2022unstructured, rao2021counterfactual, ye2021deep, he2021transreid, li2023clip} have made significant progress, they are typically domain-specific and struggle to adapt to new identities over time. 
To address this, Lifelong person ReID (LReID)~\cite{wu2021generalising, pu2021lifelong} extends the traditional person ReID task to the incremental learning scenario, allowing the model to incorporate new identities while retaining past knowledge.

A primary challenge in LReID is catastrophic forgetting, exacerbated by the differences in data distribution across cameras and environments. 
Early studies~\cite{pu2021lifelong, yu2023lifelong, ge2022lifelong} attempted to mitigate forgetting by storing and replaying old examples during training, \textit{i.e.}, rehearsal-based. However, the incremental storage of old data takes up much space, and unrestricted access to pedestrian images raises the risk of privacy breaches.  
As a result, a substantial amount of recent work~\cite{wu2021generalising, pu2022metareconciliation, xu2024lstkc, li2024exemplar} has focused on rehearsal-free technologies, using constraints or regularization to learn a ReID model shared across different tasks.

\begin{figure*}[t]
  \centering
  \includegraphics[width=1\linewidth]{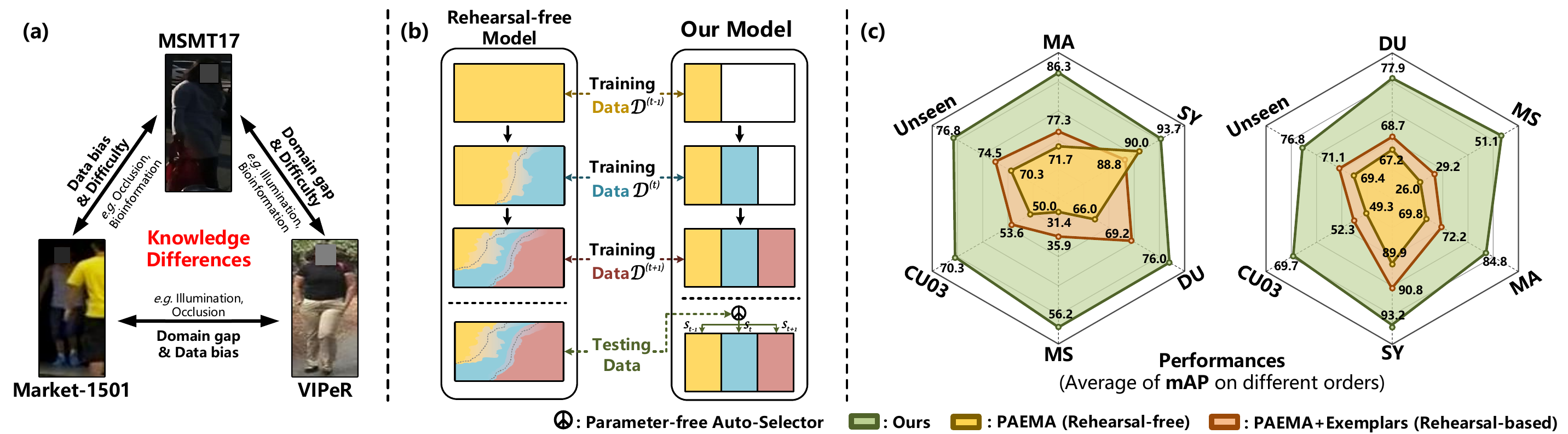}
  \caption{(a) Illustration of knowledge differences in ReID datasets. 
    (b) Key differences between our model and previous work. Typically, rehearsal-free methods incrementally regularize shared weights to balance new knowledge, which inevitably degenerates previously acquired knowledge due to learning conflict.
    By contrast, we present efficient independent modules to learn task-specific knowledge, and then dynamically tease out knowledge for the required ReID task via a parameter-free auto-selector.
    (c) Our method beats existing state-of-the-art competitors with a significant advantage in both seen and unseen domains. Please see Sec.~\ref{subsec:setup} for abbreviations of datasets.
  \label{fig:teaser}}
  \vspace{-0.15in}
\end{figure*}

In this paper, we argue that task-specific knowledge differences between tasks can cause learning conflicts in the shared model, inevitably affecting and squeezing previously acquired knowledge.  
A variety of factors, including domain gap, data bias, and task difficulty, contribute to knowledge differences.
As depicted in Fig.~\ref{fig:teaser}(a), 
images from VIPeR contain occlusion-free bias, so features learned from it may be degraded in Market-1501 with occlusions. 
Similarly, MSMT17 has significant variations in illumination, and face mosaics would erase biological information, making the ReID task more difficult than on other datasets.
Existing either rehearsal-free or rehearsal-based approaches attempt to cram different knowledge into one shared model, which cannot effectively solve the problem of learning conflict caused by knowledge differences. 
Drawing inspiration from human cognitive systems~\cite{Penfield1937somatic}, we advocate that kinds of knowledge may coexist in different containers without necessarily balancing into one. \re{A naive solution is thus to allocate distinct containers for task-specific knowledge.} 
Nevertheless, this idea faces two limitations in practice:
\textbf{(1) The storage and computational overhead must remain manageable}. In the process of lifelong learning, simply stacking knowledge containers shows incremental growth in parameters and computations, which could eventually overload the system.
\textbf{(2) The selection of knowledge containers must be dynamic and autonomous.} 
Obviously, any human intervention or using evaluation prior (\textit{e.g.}, knowing the order or names of tasks~\cite{ge2023clr}) is usually not allowed, because that will severely limit the practical applications of methods.

To address these challenges, we propose \textit{AdalReID}, a novel framework with auto-selected knowledge adapters for LReID. Inspired by advances in parameter-efficient fine-tuning~\cite{chen2022vision, jia2022visual, zhou2022learning}, we introduce knowledge adapters to learn domain-specific knowledge incrementally. 
Our method preserves past knowledge while integrating new data efficiently, overcoming catastrophic forgetting and dataset shifts.
More importantly, the parameter growth of the knowledge adapter is acceptable, and the adapters can be aggregated with a certain weight, resulting in almost no additional computational complexity during inference. 
To acquire these weights adaptively, we propose a parameter-free auto-selection strategy, which measures distribution distance across datasets as knowledge similarity to dynamically integrate compatible adapters for ReID, as illustrated in Fig.~\ref{fig:teaser}(b). 
Additionally, we develop a temperature scheduling technique to facilitate interaction between different knowledge adapters, promoting knowledge sharing at shallower layers while prioritizing domain-specific knowledge at deeper layers.
Our auto-selection strategy accelerates adaptation and is effective for unseen applications, ensuring the model's scalability and efficiency.

\noindent \textbf{Contributions.} We summarize the key contributions as follows:
\textbf{(1)} We introduce AdalReID, a framework with auto-selected knowledge adapters to address catastrophic forgetting in LReID.
\textbf{(2)} We propose a parameter-free auto-selection strategy for dynamically assigning adapters to different tasks, alongside a temperature scheduling technique for improved knowledge interaction, improving robustness on unseen tasks.
\textbf{(3)} Extensive experiments on six datasets across five lifelong learning orders demonstrate that AdalReID outperforms existing methods, achieving a new state-of-the-art performance in LReID, as shown in Fig.~\ref{fig:teaser}(c).

\section{Related Work}
\noindent \textbf{Lifelong Learning.}
Lifelong learning aims to receive and learn knowledge from new tasks continuously. Its classic issue is catastrophic forgetting, in which the model gradually forgets the old knowledge as new knowledge is introduced and integrated. Existing approaches generally tackle this problem in light of regularization, rehearsal, or architecture.
Regularization-based methods~\cite{kirkpatrick2017overcoming,li2017learning,zenke2017continual,chen2023learn} balance the old and new knowledge by constraining the update of model parameters, such as knowledge distillation and consistency loss, to prevent catastrophic forgetting.
Rehearsal-based strategies, usually paired with regularization, additionally introduce memory buffers to retain representative samples from past tasks~\cite{rebuffi2017icarl,wu2019large,hou2019learning}. While effective, these methods are limited by buffer size and data privacy risk. 
Different from the first two categories, architecture-based approaches avoid catastrophic forgetting by assigning a specific component to each new task. Components can be partially independent parameters of the model \cite{de2023effectiveness,ge2023clr,singh2020calibrating,wen2020batchensemble} or additional extended sub-networks \cite{chen2023lifelong,li2019learn,rusu2016progressive,yan2021dynamically,yang2024moral,yu2024boosting}.
In this paper, we build a structure-based framework that adopts knowledge adapters and a parameter-free routing mechanism for lifelong learning. 
\re{Nevertheless, our method differs from recent approaches~\cite{yu2024boosting,yang2024moral,chen2023lifelong} in its technical design. They typically couple a set of learnable experts with a parameterized router in a Mixture-of-Experts (MoE) style adapter, where both the experts and the routing mechanism must be jointly optimized. In contrast, we directly modularize task knowledge into independent knowledge adapters and introduce a parameter-free auto-selector to schedule them. This decoupled design allows the number of adapters to be flexibly adjusted according to deployment requirements while avoiding the training overhead of an additional routing network. Furthermore, we propose temperature scheduling and knowledge tease-out strategies to encourage effective collaboration among multiple adapters.}

\noindent \textbf{Lifelong Person Re-identification.}
Conventional person re-identification (ReID) aims to retrieve images from surveillance cameras that are consistent with the query identity~\cite{luo2020astrong,qian2020long, ye2021deep,he2021transreid}. To expand the scope of monitoring or adapting to new person identities, the lifelong person re-identification (LReID) task~\cite{wu2021generalising,pu2021lifelong} is thus proposed to learn and accumulate ReID knowledge from sequentially incoming datasets or tasks. 
Inheriting the technologies from lifelong learning, a series of regularization-based \cite{pu2021lifelong,xu2024distribution,cui2024learning,sun2022patch}, rehearsal-based \cite{yu2023lifelong,ge2022lifelong,wu2021generalising,huang2023learning,pu2023memorizing} and structure-based \cite{pu2022metareconciliation,wang2024distribution} approaches are proposed and tailored for lifelong person ReID. Concretely,
PTKP~\cite{ge2022lifelong} casts LReID as a domain adaptation problem and adopts a domain consistency loss between the old and new spaces to mitigate domain discrepancies from multiple datasets. 
KRC~\cite{yu2023lifelong} uses memory exemplars and models to refresh knowledge, enabling both forward and backward transfer in lifelong learning.
However, rehearsal-based methods incur extra storage costs and raise privacy concerns, limiting their scalability. 
In contrast, regularization-based exemplar-free methods are gaining attention. 
For example, LSTKC~\cite{xu2024lstkc} uses two modules to support short-term learning and consolidate long-term knowledge. 
MRN~\cite{pu2022metareconciliation} adapts Batchnorm layers to balance domain-dependent and independent statistics.
Recently, advanced deep architectures, such as PAEMA~\cite{li2024exemplar}, employ transformer structures and leverage visual prompting for model re-parameterization to benefit knowledge preservation, while ATFC~\cite{liu2024attribute} leverages text modality to strengthen global identity-related representations and derive attributes from text to extract local identity-free attributes for LReID. 
Nevertheless, these methods still struggle with balancing old and new knowledge, and some of them require careful tuning of optimization parameters for different lifelong learning stages.
In this paper, we propose a novel framework to dynamically route specific adapters, which further integrates the shared knowledge to customize it for different tasks. Compared with previous studies that store old task samples for knowledge reconsolidation~\cite{yu2023lifelong,huang2023learning,pu2023memorizing}, our knowledge adapters have marginal storage overhead and no risk to data privacy. Meanwhile, it doesn't require reserving any old models or classifiers for knowledge distillation~\cite{xu2024lstkc,li2024exemplar,liu2024attribute,pu2022metareconciliation}.

\section{Methodology}

\subsection{Preliminary}
\noindent \textbf{Task Formulation.} 
Given a set of person ReID datasets $\mathcal{D} = {\left\{ \mathcal{D}^{(t)} \right\}}^{T}_{t=1}$, the task of LReID aims to sequentially learn knowledge from ${\left\{ \mathcal{D}^{(t)} \right\}}^{T}_{t=1}$ in $T$ steps. At each step, the $t$-th dataset $\mathcal{D}^{(t)}=\left\{ \left(x^{(t)}_{i}, y^{(t)}_{i}\right)\right\}_{i=1}^{N^{t}}$ is available for training while samples from previous datasets $\mathcal{D}^{<t}$ are usually restricted to access considering data privacy or storage limitation. $\left(x^{(t)}_{i}, y^{(t)}_{i}\right)$ represents pairs of person images and identity labels, $N^{t}$ is the total number of images, and $C^{(t)} = \bigcup_{i=1}^{N^{(t)}} y^{(t)}_{i}$ means the set of identities which is not overlapped with other datasets, $\bigcap_{t=1}^{T}C^{(t)}=\emptyset$. At the end of lifelong learning, models are encouraged to achieve good performance on previous datasets, meanwhile, showing good generalization on unseen domains. 

\begin{figure*}
  \centering
  \includegraphics[width=1\linewidth]{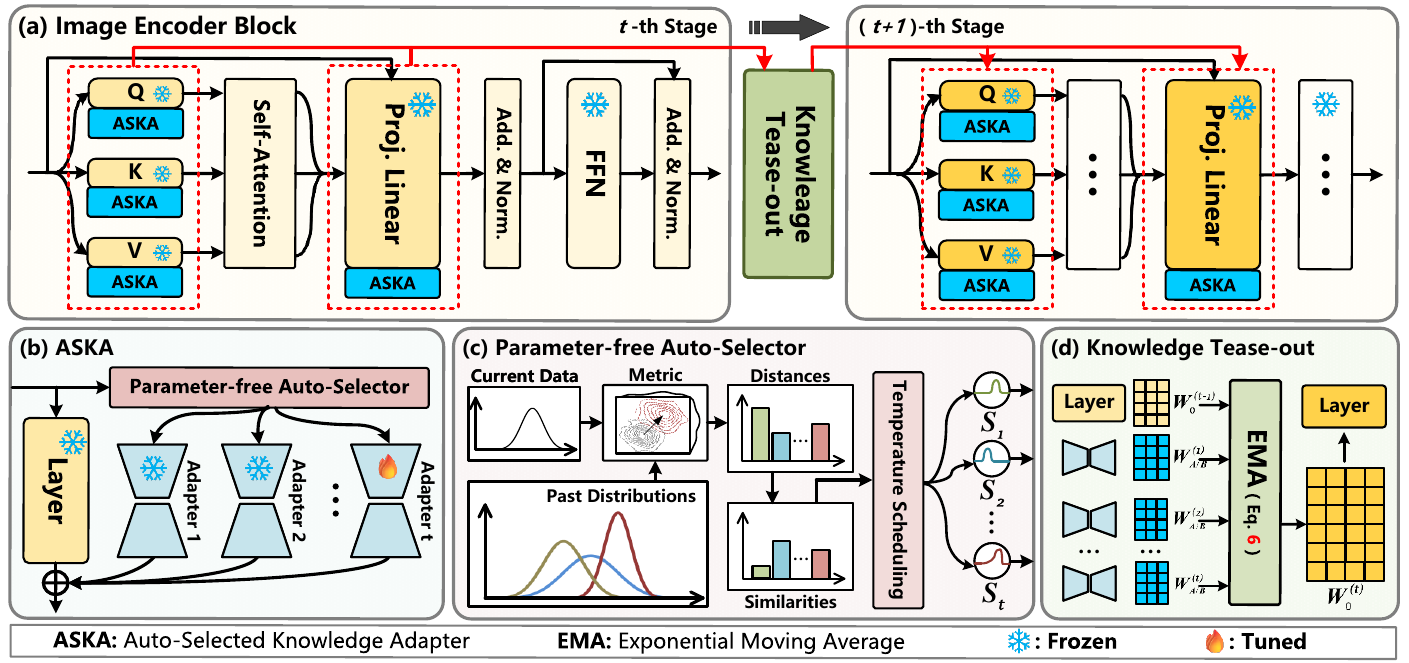}
  \caption{Overview of our framework AdalReID.
  (a) Our proposed Auto-Selected Knowledge Adapter (ASKA) is a parameter-efficient module that can be plugged into some parameterized layers of each image encoder block.
  (b) The ASKA module consists of two key components: the knowledge adapters and the knowledge auto-selector. During training, we incrementally build adapters to learn domain-specific knowledge.
  (c) Our parameter-free auto-selector estimates the knowledge similarity via the statistical distribution, which is incorporated with the temperature scheduling to adaptively select appropriate adapters for the current data or the testing set. 
  (d) At the end of each training stage, we update the shared knowledge $W_{o}$ via EMA from all adapters to strengthen its representation.
  \label{fig:framework}}
  \vspace{-0.15in}
\end{figure*}

\noindent \textbf{Method Overview.} 
Figure~\ref{fig:framework} illustrates the schematic overview of our framework, dubbed \textit{AdalReID}, that introduces auto-selected knowledge adapters for lifelong person re-identification. Our method consists of two key components, \textit{the knowledge adapter} and \textit{the knowledge auto-selector}. The former builds low-overhead adapters for each dataset to efficiently learn ReID knowledge, preventing knowledge confusion and incongruity. The latter automatically selects appropriate knowledge for input data to effectively address knowledge forgetting.
\re{In this paper, ``knowledge'' refers to the implicit identity-discriminative representation capability learned from historical person ReID datasets. This knowledge is encoded and preserved in the parameters of each knowledge adapter, whose parameters serve as its carrier rather than the knowledge itself.}

In the following sections, we first introduce parameter-efficient knowledge adapters to learn domain-specific knowledge in Sec.~\ref{subsec:adapters}. Next, we elaborate on the parameter-free knowledge selection strategy, which takes into account the domain difference and representation similarity between images (see Sec.~\ref{subsec:auto-select}). Finally, we describe 
the training and testing procedures of our AdalReID in Sec.~\ref{subsec:architecture}.

\subsection{Knowledge Accumulation via Adapters \label{subsec:adapters}}
To avoid knowledge conflicts and coverage problems, we propose to create domain-specific containers to separately learn from each incremental dataset.
Compared to previous methods \cite{yu2023lifelong,yu2023lifelong,ge2022lifelong} that aggregate knowledge from different datasets into a fixed-size model, we adopt an extensible framework to effectively preserve the old knowledge while learning the new one. However, there are two important properties that should be considered in the design of containers to accommodate the scenarios of LReID. 
First, as the number of incoming datasets continues to increase, the overhead and computation of each container must be acceptable. Second, knowledge learned by different containers is expected to be shared or complementary.

\noindent \textbf{Knowledge Adapters.} 
Inspired by the recent advances in parameter-efficient fine-tuning ~\cite{chen2022vision, jia2022visual, zhou2022learning}, we instantiate each container with the adapter \cite{hu2021lora}. Each of these is a parameter-efficient module plugged into some layer of the network. Adapters tune the model to be optimized for the specific ReID dataset. In this case, other model parameters are frozen during the fine-tuning process to protect the learned knowledge from perturbation. Concretely, we adopt LoRA \cite{hu2021lora} to further decouple the parameter matrix of the adapter into low-rank space, which can be expressed as,
\begin{equation}
\begin{small}
\tilde{f} = W_{o}\cdot f + b_{o} + \frac{\alpha}{r} W_{A}\cdot W_{B}\cdot f
\label{eq:lora}
\end{small}
\end{equation}

\noindent where $W_{o}$ and $b_{o}$ represent the original weight and bias of the parameterized layer in the model, $W_{A}$ and $W_{B}$ are two decoupled low-rank matrices with rank $r$ of the adapter, $\alpha$ is a coefficient to trade-off the importance between the original layer and the adapter. $f$ denotes the intermediate features of images. Equipped with adapters, we can comprehensively embrace new knowledge, rather than cautiously learning from datasets with a slow-paced learner \cite{yu2023lifelong,zhang2023slca}.

\noindent \textbf{Incremental Adapters.} Solely depending on one knowledge adapter still encounters catastrophic forgetting in the learning process, since the tuned weights (\textit{i.e.}, knowledge) are inevitably shifted by the intervention of subsequent datasets. To this end, we incrementally extend Eq.~\ref{eq:lora} to the form of multiple adapters, each of which is responsible for one ReID dataset $\mathcal{D}^{(t)} |_{t=1}^{T}$. At the $t$-th stage of lifelong learning, we can rewrite Eq.~\ref{eq:lora} as follows,
\begin{equation}
\begin{small}
\tilde{f} = W_{o}\cdot f + b_{o} + \frac{\alpha}{r}\sum_{t=1}^{T} s_{t} \ W^{(t)}_{A}\cdot W^{(t)}_{B}\cdot f
\label{eq:incremental_lora}
\end{small}
\end{equation}

\noindent where $s \in \mathbb{R}^{T}$ represents the similarity between knowledge stored in different adapters.
$s$ is crucial for integrating complementary knowledge and suppressing conflicting information, as explained in the next section.

\subsection{Auto-selection for Knowledge Adapters \label{subsec:auto-select}}

During the process of lifelong learning, the knowledge adapters learned at the early stage may not be aware of the existence of those built in the latter stages, and knowledge from different datasets, such as facial information in VIPeR and MSMT17\_V2, may conflict.
Therefore, instead of equally adding the outputs of adapters, we introduce $s$ to measure the similarity between knowledge in different datasets, which allows for the dynamic allocation of appropriate adapters.

\noindent \textbf{Knowledge Similarity.} In this paper, we utilize the domain similarity between datasets to refer to knowledge similarity. Specifically, at each step of lifelong learning, we additionally store the statistical distribution of the current dataset by calculating the mean $\mu$ and the covariance $\Sigma$ of image features. 
Consequently, given the $t$-th dataset $\mathcal{D}^{(t)}$ for training, we compute the domain similarity $s^{(t)} \in \mathbb{R}^{t}$ between it and all previous datasets $\mathcal{D}^{(\le t)}$  with \textit{2-Wasserstein} distance,
\begin{equation}
\begin{small}
s^{(t)}_i = \frac{exp\left(-\sqrt{d\left(\mathcal{D}^{(t)},\mathcal{D}^{(i)} \right)} / t  \right)}{ \sum_{j=1}^{t} exp\left(-\sqrt{ d\left(\mathcal{D}^{(t)},\mathcal{D}^{(j)} \right)} / t  \right)   } 
\label{eq:score}
\end{small}
\end{equation}

\begin{small}
\begin{equation}
\begin{aligned}
d\left(\mathcal{D}^{(i)},\mathcal{D}^{(j)} \right) = &|| \mu^{(i)} - \mu^{(j)} ||^{2} \\
&+ Tr\left( \Sigma^{(i)} + \Sigma^{(j)} - 2 \times \sqrt{ 
\Sigma^{(i)}\Sigma^{(j)} }     \right)
\label{eq:fid}
\end{aligned}
\end{equation}
\end{small}

\noindent where $\mu^{(i)}$ and $\Sigma^{(i)}$ are the statistical distribution of $i$-th dataset $\mathcal{D}^{(i)}$, $Tr\left(\cdot\right)$ means the trace of the input matrix, and $t$ is the temperature. Note that we do not revisit or store previous images, thus adhering to the requirements of lifelong learning. Additionally, storing $\mu$ and $\Sigma$ only costs limited storage resources (see our discussions in Tab.~\ref{tab:ablation_cost} and \ref{tab:measure}). During training, we split 15\% of the training set $\mathcal{D}^{(t)}$ as the validation set to perform Eq.~\ref{eq:score} when $i = t$. And we directly use the testing set to calculate its knowledge similarity to all learned datasets $\mathcal{D}^{(t)} | _{t=1}^{T}$ for inference. 

\noindent \textbf{Temperature Scheduling.} 
One special case of $s$ is a one-shot vector, in which case every knowledge adapter is independent of the others. Nevertheless, we argue that the purpose of lifelong learning is to achieve the complementarity of multi-domain knowledge through incremental learning, improving the model performance as well as its generalization ability. Inspired by multi-task learning that similar tasks could share low-level features~\cite{ando2005framework, zhang2021survey, su2017multi, chen2009convex}, we thus present another strategy of temperature scheduling, 
\begin{equation}
\begin{small}
t = \delta\left( l, g\right) = a\times g\left(\frac{l}{L}\right) + b
\label{eq:schedule}
\end{small}
\end{equation}

\noindent where $l$ and $L$ indicate the number of current layers and the total number of layers in the model, respectively. $a$ and $b$ mean scaling and shifting factors. $g\left(\cdot\right)$ is a monotonically decreasing function, like \textit{linear}, \textit{concave} or \textit{convex}. Intuitively, the deeper the model layer the knowledge adapter is inserted into, the smaller the temperature coefficient, resulting in a steeper distance distribution from Eq.~\ref{eq:score}. Models are encouraged to incorporate knowledge from different adapters at the shallow-level layers, while at the deeper layers, they make more use of domain-specific knowledge.

\noindent \textbf{Knowledge Tease-out.} Our proposed strategy automatically selects appropriate knowledge adapters for different testing datasets, with no need for any testing prior, such as the order of testing datasets or their names \cite{chen2023lifelong,ge2023clr}.
More importantly, we can merge the weights of different adapters to tease out the required knowledge. Concretely, after the $t$-th stage of training, we update the weight of $W_{o}^{(t-1)}$ via an exponential moving average from all adapters to strengthen the representation of the shared knowledge,
\begin{equation}
\begin{small}
\begin{aligned}
{W}_{o}^{(t)} &= \beta{W}_{o}^{(t-1)} + \left(1 - \beta\right) \left({W}_{o}^{(t-1)} + \frac{\alpha}{r \times t}\sum_{i=1}^{t} \ W^{(i)}_{A}\cdot W^{(i)}_{B} \right) \\
&= {W}_{o}^{(t-1)} + \left(1 - \beta\right) \frac{\alpha}{r \times t}\sum_{t=1}^{t} \ W^{(i)}_{A}\cdot W^{(i)}_{B}
\label{eq:ema}
\end{aligned}
\end{small}
\end{equation}

\noindent where ${W}_{o}^{(0)} = {W}_{o}$. 
\re{Interestingly, our framework provides additional flexibility for deployment under strict memory constraints. We can selectively remove specific historical adapters according to deployment requirements, trading task-specific knowledge for additional storage savings. 
Furthermore, for a specific deployment, we could integrate all learned $T$ adapters into the backbone based on similarities $s$, \textit{i.e.}, $\tilde{W}_{o} = W_{o}^{(T)} + \frac{\alpha}{r}\sum_{t=1}^{T} s_{t} \ W^{(t)}_{A}\cdot W^{(t)}_{B}$. After merging, inference can be performed using only the updated backbone, thereby introducing almost no additional computational overhead from adapters. If memory capacity is even more limited, the merged backbone $\tilde{W}_{o}$ can further serve as the initialization for the next lifelong learning cycle, allowing historical adapters to be discarded while retaining the accumulated knowledge.}

\subsection{Details of Training and Inference\label{subsec:architecture}}

The AdalReID framework is built on the CLIP-ReID backbone \cite{li2023clip}, leveraging image-text multimodality to learn robust feature representations. Specifically, AdalReID includes CLIP's pre-trained image encoder $\mathcal{F}^{(t)}$ and text encoder $\mathcal{T}$. 
We implement the proposed auto-selected knowledge adapters in the image encoder $\mathcal{F}^{(t)}$ and incrementally train it while keeping $\mathcal{T}$ frozen.
$\mathcal{F}^{(t)}$ and $\mathcal{T}$ alternatively learn and transfer knowledge distilled from the incoming ReID dataset $\mathcal{D}^{(t)}$ in the image and text space.
   
\noindent \textbf{Text side.} 
The text encoder $\mathcal{T}$ takes customized prompt of ``$A \ photo \ of \ a \ [{X}_{1}][{X}_{2}] \cdots [{X}_{|C^{(t)}|}] \ person.$'' as input, where $[{X}_{i}]$ is the learnable identity token for each identity in $\mathcal{D}^{(t)}$ and $|\cdot|$ denotes cardinality. The image encoder $\mathcal{F}^{(t-1)}$ extracts images features from $\mathcal{D}^{(t)}$ to supervise the learning of $[{X}_{i}]$ via the contrastive loss $\mathcal{L}_{t2i}$ and $\mathcal{L}_{i2t}$ \footnote{We refer the readers to \cite{li2023clip} for details of training losses. \label{fn:loss}}.

\noindent \textbf{Image side.} 
After optimizing $[{X}{i}]$ at the $t$-th step, we freeze the identity tokens on the text side and train the newly added $t$-th adapter in the image encoder $\mathcal{F}^{(t)}$. 
The objective consists of two parts: 
the first part reuses text features from $\mathcal{T}$ to compute the image-to-text cross-entropy loss $\mathcal{L}_{t2tce}$ \footref{fn:loss}, and the second part includes the conventional triplet loss $\mathcal{L}{tri}$ \cite{hermans2017defense} and ID loss $\mathcal{L}{id}$ \cite{luo2019bag}.

By regarding the text space as the intermediary, knowledge from different ReID domains can be effectively transferred and accumulated via $\mathcal{F}^{(0)} \to \mathcal{T} \to \mathcal{F}^{(1)} \to \mathcal{T} \to \cdots \to \mathcal{F}^{(T)}$, where $\mathcal{F}^{(0)}$ means the initial image encoder with CLIP pre-trained weights. After training $T$ steps, we utilize the image encoder $\mathcal{F}^{(T)}$ to extract features of testing images from $\mathcal{D}^{t}|_{t=1}^{T}$.
The cosine distances between the query and gallery images are computed as metrics for ReID.

\begin{table}[t]
    \centering
    \footnotesize
    \caption{\textbf{Statistics of ReID datasets for lifelong learning.} 
    ``Trials'' refers to the number of different training–testing splits used in the experiments. Because some of them are randomly partitioned, multiple trials are conducted to ensure the reliability and stability of the results.
    \label{tab:dataset}}
    \vspace{-0.05in}
    \setlength{\tabcolsep}{0.9mm}{
    \begin{tabular}{llccccccc}
        \toprule
        & \multirow{2}{*}{\textsc{Datasets}} & \multicolumn{3}{c}{\textsc{Original Identities}} & \multirow{2}{*}{\textsc{Trials}}\\
        \cmidrule(r){3-3} \cmidrule(r){4-5} 
        & & \textbf{Train} & \textbf{Query} & \textbf{Gallery} & \\
        \midrule
        \multirow{5}{*}{\textsc{Seen}} & Market-1501~\cite{zheng2015scalable} & 751 & 750 & 751 & 1 \\
        & MSMT17\_V1/V2~\cite{wei2018person} & 1041 & 3060 & 3060 & 1  \\
        & DukeMTMC-ReID~\cite{zheng2017unlabeled}  & 702 & 702 & 1110 & 1  \\
         & LPW~\cite{song2018region}  & 875 & 876 & 876 & 1  \\
        & CUHK03-NP~\cite{zhong2017re} & 767 & 700 & 700 & 1 \\
        & CUHK-SYSU~\cite{xiao2016endtoend} & 942 & 2900 & 2900 & 1  \\
        & VIPeR~\cite{gray2008viewpoint}  & 316 & 316 & 316 & 1\\
        \midrule
        \multirow{8}{*}{\textsc{Unseen}} & VIPeR~\cite{gray2008viewpoint}  & 316 & 316 & 316 & 20  \\
        & PRID~\cite{hirzer2011person} & 100 & 100 & 649 & 10 \\
        & GRID~\cite{loy2010time}  & 125 & 125 & 126 & 10  \\
        & i-LIDS~\cite{zheng2009associating} & 43 & 60 & 60 & 10  \\
        & CUHK01~\cite{li2013human} & 485 & 486 & 486 & 10  \\
        & CUHK02~\cite{li2013locally} & 1677 & 239 & 239 & 1 \\
        & CUHK03~\cite{li2014deepreid} & 767 & 700 & 700 & 20 \\
        & SenseReID~\cite{zhao2017spindle} & 1718 & 521 & 1718 & 1 \\
        \bottomrule
    \end{tabular}}
\vspace{-0.2in}
\end{table}

\section{Experiments}
\subsection{Experimental Setup \label{subsec:setup}}
\subsubsection{Dataset}
Following previous literature \cite{pu2021lifelong,yu2023lifelong,ge2022lifelong,huang2023learning,pu2023memorizing}, \re{we perform lifelong learning on 13 ReID datasets\footnote{Datasets are downloaded from the projects of \href{https://kaiyangzhou.github.io/deep-person-reid/datasets.html}{torchreid} and \href{https://github.com/BJTUJia/person_reID_DualNorm}{DualNorm}.}, including Market-1501~\cite{zheng2015scalable}, MSMT17~\cite{wei2018person}, DukeMTMC-reID~\cite{zheng2017unlabeled}, LPW~\cite{song2018region}}， CUHK-SYSU~\cite{xiao2016endtoend}, CUHK03~\cite{li2014deepreid}, VIPeR~\cite{gray2008viewpoint}, PRID~\cite{hirzer2011person}, GRID~\cite{loy2010time}, i-LIDS~\cite{zheng2009associating}, CUHK01~\cite{li2013human}, CUHK02~\cite{li2013locally} and SenseReID~\cite{zhao2017spindle}. 
These datasets are divided into two groups of \textbf{Seen} and \textbf{Unseen} domains for standard supervised learning and generalization evaluation, respectively. Particularly, the VIPeR dataset is selected for both domains. The original CUHK03 (detected) dataset is used for the unseen domain, and its variant of CUHK03-NP~\cite{zhong2017re} (detected) is for the seen domain. 
The statistics of ReID datasets are shown in Tab.~\ref{tab:dataset}. We will elaborate on the training and testing protocols of LReID below. 

\noindent \textbf{Training:} 
For a fair comparison, we take the same training orders and protocols as existing LReID studies~\cite {wu2021generalising,ge2022lifelong,pu2022metareconciliation,pu2021lifelong}. \re{In a nutshell, there are seven popular LReID orders in the community, as illustrated in Tab.~\ref{tab:seen-orders}}. 
We sequentially train the model from the first task (dataset) to the last for each training order.
Following MEGE~\cite{pu2023memorizing}, \re{we uniformly sample 500 identities from each dataset in Orders-1, 2, 6, and 7 to build the balanced protocol.}
\textit{Notably, we keep the same hyperparameters and optimization parameters for all protocols and experiments, without using any prior (\textit{e.g.}, adjusting a few parameters according to the order or names of datasets).}

\noindent \textbf{Testing:}
After the last lifelong learning stage, we utilize the image encoder to extract features of the corresponding testing sets from both seen and unseen datasets. The cosine distances between the query and gallery images are computed as metrics for ReID. In addition, since the query and gallery splitting of the earlier datasets (\textit{e.g.}, VIPeR, PRID, GRID, i-LIDS, CUHK01, and CUHK03) are not deterministic, we produce multiple trials to eliminate performance fluctuations due to randomness\footnote{\url{https://github.com/TPCD/LifelongReID}}. The number of trials for each dataset is listed in Tab.~\ref{tab:dataset}, and we report the average results of all trials. Among them, we adopt the first split of the VIPeR dataset when it is used as one of the seen datasets.

\begin{table*}[t]
  \centering
  \footnotesize
  \caption{\textbf{Seven popular arrangements of training orders in lifelong person ReID.} 
  Models are trained sequentially from datasets 1 to 5, and evaluated on the corresponding testing sets after the last training stage. We use the detected set of CUHK03-NP. For unseen domain evaluation, the order numbers correspond to those in Tab.~\ref{tab:unseen-orders}.
  }
   \vspace{-0.05in}
  \label{tab:seen-orders}
    \setlength{\tabcolsep}{4.5mm}{
    \begin{tabular}{lcccccc}
    \toprule
    \multicolumn{1}{c}{\multirow{2}[0]{*}{\textsc{\#Order}}} & \multirow{2}[0]{*}{\textsc{Year}} & \multicolumn{5}{c}{\textsc{Seen Datasets}}  \\  
    \cmidrule(r){3-7} 
   & & 1 & 2 & 3 & 4 & 5 \\
    \midrule
   1~\cite{pu2021lifelong} & 2021 & Market-1501 & CUHK-SYSU & DukeMTMC-ReID & MSMT17\_V2 & CUHK03-NP \\
   2~\cite{pu2021lifelong} & 2021 & DukeMTMC-ReID & MSMT17\_V2 & Market-1501 & CUHK-SYSU & CUHK03-NP\\
   3~\cite{wu2021generalising} & 2021 & Market-1501 & DukeMTMC-ReID & CUHK-SYSU & MSMT17\_V1 & - \\
   4~\cite{pu2022metareconciliation} & 2022 & Market-1501 & CUHK-SYSU & MSMT17\_V2 & CUHK03-NP & -  \\
   5~\cite{yu2023lifelong} & 2023 & VIPeR & Market-1501 & CUHK-SYSU & MSMT17\_V2 & - \\
    6~\cite{xu2026vision} & 2026 & Market-1501 & CUHK-SYSU & LPW & MSMT17\_V2 & CUHK03-NP \\
    7~\cite{xu2026vision} & 2026 & LPW & MSMT17\_V2 & Market-1501 & CUHK-SYSU & CUHK03-NP\\
    \bottomrule
    \end{tabular}
    }
 \vspace{-0.05in}
\end{table*}

\begin{table*}[t]
  \centering
  \footnotesize
  \caption{\textbf{Seven popular arrangements of unseen datasets in lifelong person ReID.}
  Three studies~\cite{wu2021generalising, ge2022lifelong, huang2023learning} employ different unseen datasets for generalization evaluation, we thus select the common CUHK03 dataset among these studies as the unseen dataset. We use the detected set of CUHK03. For training, the order numbers correspond to those in Tab.~\ref{tab:seen-orders}.}
   \vspace{-0.05in}
  \label{tab:unseen-orders}
    \setlength{\tabcolsep}{4mm}{
    \begin{tabular}{lccccccccc}
    \toprule
\multicolumn{1}{c}{\multirow{2}[0]{*}{\textsc{\#Order}}} & \multirow{2}[0]{*}{\textsc{Year}} & \multicolumn{8}{c}{\textsc{Unseen Datasets }}   \\  
    \cmidrule(r){3-10}
    & & VIPeR & PRID & GRID & i-LIDS & CUHK01 & CUHK02 & CUHK03 & SenseReID \\  
    \midrule
   1~\cite{pu2021lifelong} & 2021 & $\checkmark$ & $\checkmark$ & $\checkmark$ & $\checkmark$ & $\checkmark$ & $\checkmark$ & & $\checkmark$  \\
   2~\cite{pu2021lifelong} & 2021 & $\checkmark$ & $\checkmark$ & $\checkmark$ & $\checkmark$ & $\checkmark$ & $\checkmark$ & & $\checkmark$ \\
   3~\cite{wu2021generalising,ge2022lifelong,huang2023learning} & 2021 & & & & & & & $\checkmark$ & \\
   4~\cite{pu2022metareconciliation} & 2022 & $\checkmark$ & $\checkmark$ & $\checkmark$ & $\checkmark$ & $\checkmark$ & $\checkmark$ & & $\checkmark$ \\
   5~\cite{yu2023lifelong} & 2023 & & $\checkmark$ & & & $\checkmark$ & $\checkmark$ & $\checkmark$ & \\
    6~\cite{xu2026vision} & 2026 & $\checkmark$ & $\checkmark$ & $\checkmark$ & $\checkmark$ & $\checkmark$ & $\checkmark$ & & $\checkmark$  \\
    7~\cite{xu2026vision} & 2026 & $\checkmark$ & $\checkmark$ & $\checkmark$ & $\checkmark$ & $\checkmark$ & $\checkmark$ & & $\checkmark$ \\
    \bottomrule
    \end{tabular}
    }
 \vspace{-0.2in}
\end{table*}

\subsubsection{Evaluation Metrics} 
\re{Evaluation is performed after the last step of lifelong learning. 
Following lifelong person ReID evaluation protocols~\cite{wang2025distribution,pu2021lifelong}, we adopt mean Average Precision (mAP) and Rank-1 accuracy (R-1) as metrics. To verify the quality of knowledge accumulation, we also report average accuracy by averaging mAP and Rank-1 across all datasets. 
The Average Precision (AP) evaluates the retrieval performance by considering the precision at the positions of correctly retrieved samples. It is defined as,
\begin{equation}
AP_i=\frac{1}{N_{gt}^{i}}\sum_{k=1}^{N}P_i(k)\cdot rel_i(k)
\end{equation}
where $N$ denotes the number of gallery images, $N_{gt}^{i}$ represents the number of ground-truth matches for the $i$-th query image, $P_i(k)$ denotes the precision at the $k$-th ranked position, and $rel_i(k)$ is an indicator function that equals 1 if the $k$-th ranked gallery image is a correct match and 0 otherwise. 
The mean Average Precision (mAP) is then calculated by averaging the AP values over all query images
\begin{equation}
mAP=\frac{1}{N_q}\sum_{i=1}^{N_q}AP_i
\end{equation}
where $N_q$ denotes the number of query images.}

\re{Rank-1 accuracy (R-1) measures whether the top-ranked gallery image correctly matches the identity of the query image, which can be formulated as:
\begin{equation}
Rank\text{-}1=\frac{1}{N_q}\sum_{i=1}^{N_q}\mathbb{I}(r_i^1=y_i),
\end{equation}
where $r_i^1$ denotes the identity label of the top-ranked gallery image for the $i$-th query image, $y_i$ represents the identity label of the query image, and $\mathbb{I}(\cdot)$ is the indicator function.}

\begin{figure*}[t]
  \centering
  \includegraphics[width=1\linewidth]{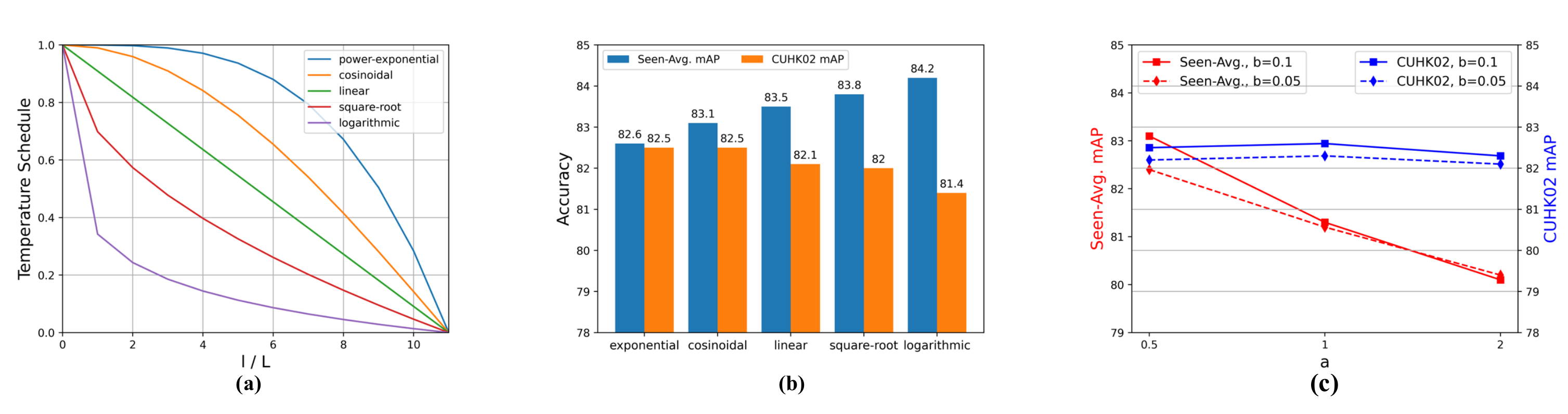}
  \vspace{-0.25in}
  \caption{\textbf{Choices of temperature scheduling}. 
  \textbf{(a)} We visualize choices of monotonically decreasing function $g(\cdot)$.
  \textbf{(b)} mAP performance of using different functions ($a=0.5$, $b=0.1$).
  \textbf{(c)} Given \textit{cosinoidal} function, we show mAP of varying $a$ and $b$. 
  }
  \label{fig:temp_schedule}
 \vspace{-0.15in}
\end{figure*}

\subsubsection{Implementation Details} 
Our framework is implemented on the PyTorch platform using 1 NVIDIA A10 GPU.
We use CLIP-ReID~\cite{li2023clip} with a pre-trained ViT-B/16 backbone. Knowledge adapters are integrated into the Query, Key, Value, and Projection layers of each image branch block, with a rank of $r=64$ and $\alpha=256$. 
To promote knowledge interaction between adapters, we apply a cosinusoidal function $g\left(\cdot\right)$ with $a=0.5$, $b=0.1$, and $\beta=0.5$. 
All person images are resized to $256 \times 128$. 
The text and image sides are trained alternately for 120 and 60 epochs, with learning rates of $3.5e^{-4}$ and $2.5e^{-5}$, respectively. The training time is about 6 hours.
The batch size is 64. 
Image features are extracted using Inception-V3 \cite{inception_v3} with 768 dimensions for statistical distribution.

\subsubsection{Implementation of Competitors}
Many recent person ReID efforts have embraced the advantages of transformer or multi-modal architectures, achieving more advanced progress. 
However, previous lifelong person ReID approaches still principally use ResNet-50 as the backbone. 
Our proposed auto-selected knowledge adapter is designed to build upon CLIP-ReID, which is a CLIP-based backbone for ReID tasks. To allay any concerns, we select several representative works and also reproduce them with the CLIP backbone to verify the effectiveness of our method. These works follow the priority criteria, such as whether they are SOTA methods in LReID, whether they have similar motivation in conventional lifelong learning tasks, whether they release source code, and so on. All methods marked with `ViT-B/16 (CLIP)' strictly adopt both image and text branches of CLIP. Here, we list all codebases used in our paper.

\begin{itemize}
    \item LSTKC~\cite{xu2024lstkc}:\url{https://github.com/zhoujiahuan1991/AAAI2024-LSTKC}
    \item DKP~\cite{xu2024distribution}: \url{https://github.com/zhoujiahuan1991/CVPR2024-DKP}
    \item PAEMA~\cite{li2024exemplar}: \url{https://github.com/zhoujiahuan1991/IJCV2024-PAEMA}
    \item MoRAL~\cite{yang2024moral}: \url{https://arxiv.org/pdf/2402.11260}
\end{itemize}

Among them, we reproduce MoRAL with technical details described in its paper. Specifically, we build the experts and router network upon CLIP-ReID, the same backbone as ours. In strict adherence to the original MoRAL paper, we incorporated the MoRAL module, consisting of eight experts, into each FNN layer. The hyperparameter top-$k$ for the router network is set to 2, aligning with its original implementation. 

\begin{table}[t]
	\begin{minipage}[t]{0.47\linewidth}
		\centering
            \footnotesize
		\caption{Different choices of temperature $t$ in Eq.~\ref{eq:schedule}. \label{tab:ablation_temp}}
		\vspace{-0.05in}
		\setlength{\tabcolsep}{0.4mm}{
		\begin{tabular}{lcccc}
            \toprule
            \multicolumn{1}{c}{\multirow{2}[0]{*}{$t$}} & \multicolumn{2}{c}{\textbf{\textsc{Seen-Avg.}}} & \multicolumn{2}{c}{\textbf{\textsc{CUHK02}}}   \\
            \cmidrule(r){2-3} \cmidrule(r){4-5}
            & mAP & R-1 & mAP & R-1 \\
            \midrule
            one-hot ($t=0.05$) & 84.2 & 92.2 & 79.6 & 79.7 \\
            uniform ($t=2.0$)  & 78.2 & 89.0 & 81.3 & 80.8 \\
            learnable & 79.4 & 89.3 & 81.9 & 81.0 \\
            \midrule
            Ours & 83.1 & 91.4 & 82.5 & 81.4 \\
            
		\bottomrule 
		\end{tabular}}
	\end{minipage}
	\hspace{0.55in}
	\begin{minipage}[t]{0.38\linewidth}
		\centering
             \footnotesize
		\caption{Storage cost of different modules. 
        \label{tab:ablation_cost}}
        \vspace{-0.05in}
		\setlength{\tabcolsep}{3mm}{
		\begin{tabular}{lc}
            \toprule
            \multicolumn{1}{c}{\multirow{2}[0]{*}{\textbf{Module}}} & \textbf{Storage} \\
             & \textbf{(MB)} \\
            \midrule
            Backbone & 474.1 \\
            \midrule
            Adapter & 18.0 \\
            $\mu$ and $\Sigma$ & 2.3 \\
            \midrule
            Exemplar & 187.5 \\
		\bottomrule 
		\end{tabular}}
	\end{minipage}
\vspace{-0.08in}
\end{table}

\begin{table}[t]
\centering{
\footnotesize{
\setlength{\tabcolsep}{2.3mm}
\caption{
Choices of measuring distribution distance.
\label{tab:measure}}
\vspace{-0.05in}
\begin{tabular}{lcccc}
    \toprule
    \multicolumn{1}{c}{\multirow{2}[0]{*}{\textbf{\textsc{Measure}}}} & \multicolumn{2}{c}{\textbf{\textsc{Seen-Avg.}}} & \multicolumn{2}{c}{\textbf{\textsc{CUHK02}}}   \\
    \cmidrule{2-5}
    & mAP & R-1 & mAP & R-1 \\
    \midrule
    Maximum Mean Discrepancy (MMD) & 83.0 & 91.3 & 81.9 & 80.0 \\
    Kernel Inception Distance (KID) & 83.2 & 91.4 & 82.3 & 80.7 \\
    2-Wasserstein distance & 83.1 & 91.4 & 82.5 & 81.4 \\
    \midrule
     Random Selection & 69.7 & 83.6 & 80.7 & 79.4 \\
     Latest Selection & 71.4 & 83.1 & 80.0 & 78.5 \\
    \bottomrule
    \end{tabular}}}
\vspace{-0.2in}
\end{table}

\subsection{Ablation Studies \label{subsec:ablation}}

We first provide in-depth studies to dissect the efficacy of different designs in our framework. 
For efficiency, we choose the four-stage training order for ablation studies instead of five. Without loss of generality,
experiments in this section follow the training order of \textbf{Market-1501$\to$ DukeMTMC-reID$\to$ CUHK-SYSU $\to$ MSMT17}, and unseen evaluation is performed on \textbf{CUHK02}. 
``$\beta$'' is set to 1.0 for simplicity.

\subsubsection{Lifelong Learning or Fine-tuning?}
As discussed in Sec.~\ref{subsec:auto-select}, a special case of the knowledge similarity $s$ is a one-hot vector, in which no interaction occurs between knowledge adapters. It unexpectedly turns lifelong learning into fine-tuning, in which the model is efficiently tuned on one dataset and evaluated on the corresponding testing set. Intuitively, fine-tuning can fit a model into one dataset to achieve better performance, but it may defeat the purpose of lifelong learning, \textit{i.e.}, improving generalization through incremental learning. To verify this, we manually set the temperature $t=0.05$, producing a one-hot similarity distribution (\textit{i.e.}, 0.99 \textit{v.s.} others). 
Compared with our full model using the temperature scheduling of Eq.~\ref{eq:schedule}, `one-hot' improves only 1.1\%/0.8\% of mAP/Rank-1 on seen domains, but drops 2.9\% and 1.7\% on the unseen CUHK02 dataset in Tab.~\ref{tab:ablation_temp}. It demonstrates the advantage of lifelong learning over fine-tuning in generalization ability and supports the motivation of our proposed temperature scheduling strategy.

To further investigate the efficacy of our temperature scheduling, we conduct two variants by uniformly merging knowledge from all adapters (implemented by setting $t=2.0$) or setting $t$ as a learnable parameter (`Learnable' for short). Both of them get inferior performance to ours. We explain that simply mixing different knowledge may lead to inconsistent information, and the self-learning of $t$ may be sub-optimal or difficult due to the domain gap. It effectively suggests the benefits of our proposed temperature scheduling in terms of both performance and generalization.

\subsubsection{Memory of Knowledge Adapters}
One concern is whether the gradual increase in the number of knowledge adapters over the process of lifelong learning will result in significant storage overhead. To answer it, we provide an analysis of the storage consumption of different components in Tab.~\ref{tab:ablation_cost}.
Existing state-of-the-art methods~\cite{ge2022lifelong,yu2023lifelong,wu2021generalising,huang2023learning} draw support from exemplars to constrain models from forgetting past knowledge. They usually keep 500 images of 250 identities at each training step. Assuming the size of each image is $256 \times 128 \times 3$, the overhead to store these images is about 180 MB for each dataset. 
By contrast, the additional storage consumption introduced by our knowledge adapters at each training step is about 18 MB with a matrix rank of $64$. Besides, it costs about 2.3 MB to save the statistical distribution of $\mu \in \mathbb{R}^{768}$ and $\Sigma \in \mathbb{R}^{768\times 768}$ per dataset, which is much cheaper. Not to mention that we don't need to keep old models or classifiers for knowledge distillation like \cite{pu2021lifelong,huang2023learning,wu2021generalising,pu2022metareconciliation,xu2024lstkc,li2024exemplar}. This indicates the practicability and extensibility of our approach. 

\begin{table*}[t]
  \centering
  \footnotesize
  \caption{\textbf{Ablation studies on different layers the knowledge adapters are plugged into.} `Q', `K', `V', and `Proj' refer to the Query, Key, Value, and Projection in the multi-head attention layer. `FFN' means feed-forward network. $\beta=1$, and we set $r=64$ and $\alpha=256$ by default to build adapters.}
   \vspace{-0.05in}
  \label{tab:compare_layer}
    \setlength{\tabcolsep}{5.2mm}{
    \begin{tabular}{ccccccccccc}
    \toprule
    \multirow{2}[0]{*}{\textsc{No.}} & \multicolumn{5}{c}{\textbf{\textsc{Layers}}}  & \multicolumn{2}{c}{\textbf{\textsc{Seen-Avg.}}} & \multicolumn{2}{c}{\textbf{\textsc{CUHK02}}} & \multirow{2}[0]{*}{\textbf{\textsc{Storage (MB)}}}   \\  
    \cmidrule(r){2-6} \cmidrule(r){7-8} \cmidrule(r){9-10}
   & Q & K & V & Proj & FFN & mAP & R-1 & mAP & R-1 &  \\
    \midrule
   1 & $\checkmark$ & & $\checkmark$ &  & & 82.0 & 90.2 & 82.0 & 79.1 & 9.0 \\
   2 & $\checkmark$ & $\checkmark$ & $\checkmark$ &  & & 82.3 & 90.6 & 81.6 & 78.1 & 13.5\\
   \midrule
   3 & $\checkmark$ &  & $\checkmark$ & $\checkmark$ & & 82.9 & 91.2 & 82.8 & 81.4 & 13.5 \\
   4 & $\checkmark$ & $\checkmark$ & $\checkmark$ & $\checkmark$ & &  83.1 & 91.4 & 82.5 & 81.4 & 18.0 \\
   \midrule
   5 &  &  &  &  & $\checkmark$ & 83.3 & 91.6 & 82.1 & 79.7 & 22.5 \\
   6 & $\checkmark$ & $\checkmark$  & $\checkmark$  & $\checkmark$  & $\checkmark$ & 84.7 & 92.2 & 83.8 & 82.0 & 40.5 \\
    \bottomrule
    \end{tabular}}
 \vspace{-0.1in}
\end{table*}

\begin{figure*}[t]
\begin{minipage}[t]{0.32\linewidth}
  \centering
  \includegraphics[width=0.9\linewidth]{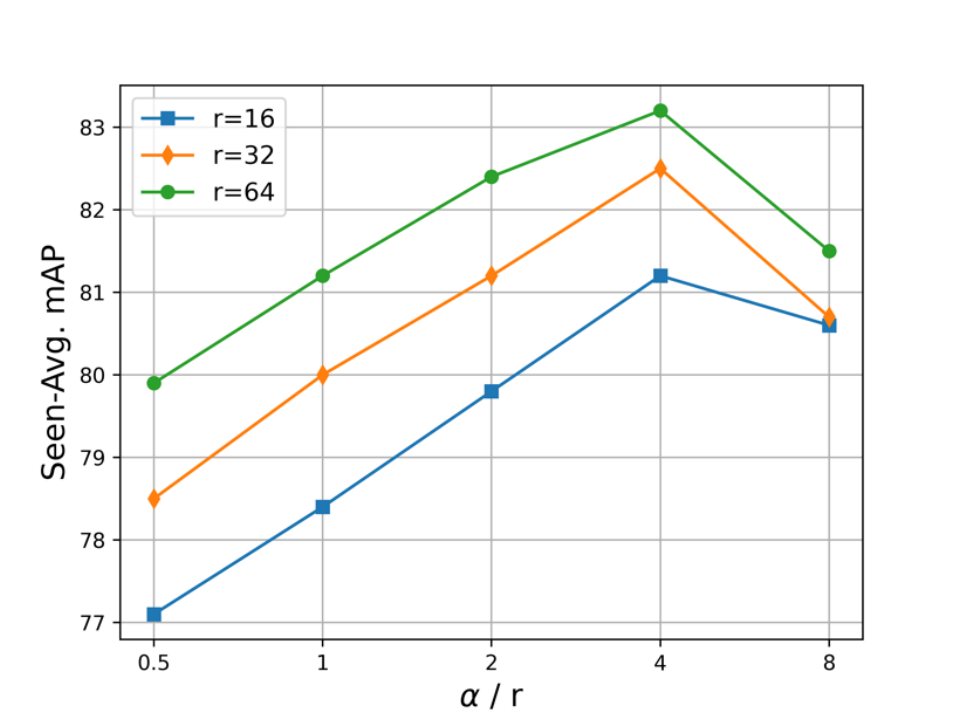}
  \vspace{-0.05in}
  \caption{Results on seen domains by varying different values of $r$ and $\alpha$ in Eq.~\ref{eq:incremental_lora}.
  }
  \label{fig:ablation_rank}
\end{minipage} 
\hspace{0.02in}
\begin{minipage}[t]{0.32\linewidth}
    \centering
    \includegraphics[width=0.9\linewidth]{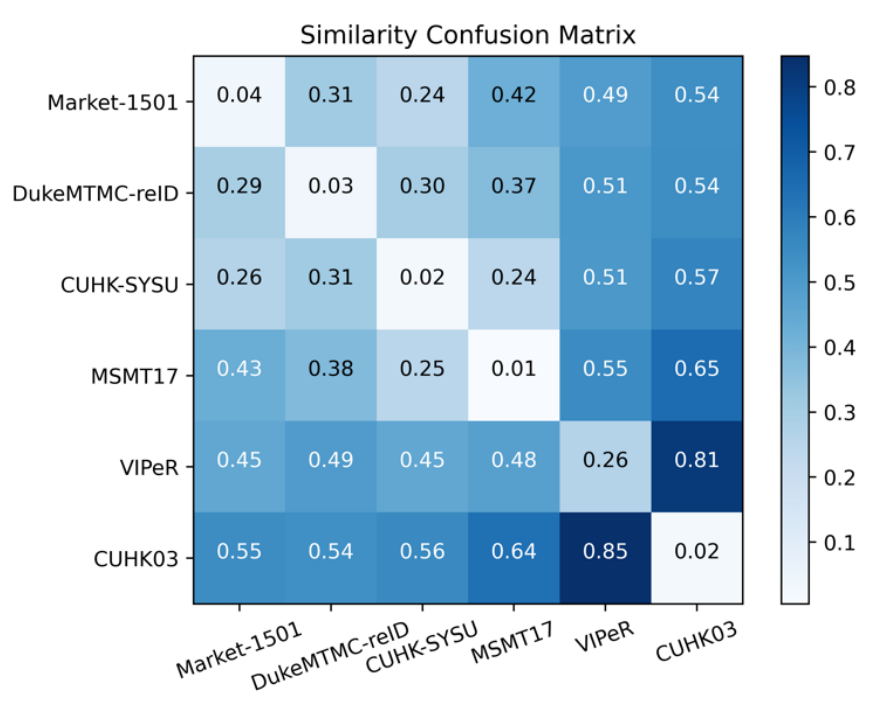}
    \vspace{-0.05in}
  \caption{Confusion matrix of knowledge similarity between ReID datasets.}
  \label{fig:ablation_cm}
\end{minipage}
\hspace{0.02in}
\begin{minipage}[t]{0.32\linewidth}
    \centering
    \includegraphics[width=0.9\linewidth]{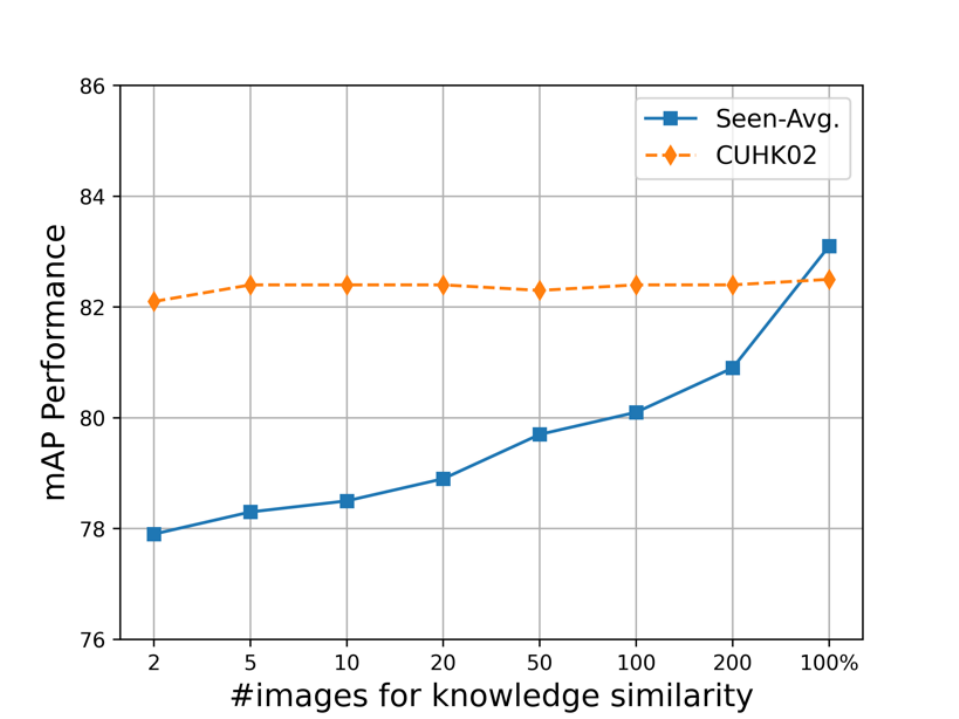}
    \vspace{-0.05in}
  \caption{Impact of \#images used to calculate knowledge similarity.
  }
  \label{fig:ablation_sample}
\end{minipage}
\vspace{-0.2in}
\end{figure*}

\subsubsection{Choices of Temperature Scheduling}
We further investigate the effects of different temperature scheduling strategies on performance and generalization of lifelong learning. 
Firstly, we visualize five different functions of $g\left(\cdot\right)$ in Fig.~\ref{fig:temp_schedule}(a), including two concave functions of \textit{exponential} and \textit{cosinoidal}, two convex functions of \textit{logarithmic} and \textit{square-root}, and one linear function. The concave functions encourage adapters to contribute more knowledge at the shallow layers, while the convex functions, in contrast, focus more on the independence between adapters.
Next, we report the results of mAP on both seen and unseen domains when using the aforesaid functions to compute the temperature $t$ in Eq.~\ref{eq:schedule}. The values of $a$ and $b$ are set to $0.5$ and $0.1$ by default to control variables. Results in Fig.~\ref{fig:temp_schedule}(b) show that, from concave to convex functions, the performance on seen datasets gradually increases, but it sacrifices the capability on the unseen domain. 
As expected, convex functions make $s$ of most network layers a one-hot vector, but apparently, the unseen domain does not belong to any seen domain.
To balance it, we adopt the \textit{cosinodial} function as $g(\cdot)$ for all experiments in Sec.~\ref{subsec:sota}.
Lastly, we study the importance of hyperparameters $a$ and $b$ in Eq.~\ref{eq:schedule}. Figure~\ref{fig:temp_schedule}(c) shows that the performance on the seen and unseen domains has an opposite trend with the increase of $a$. The choice after the trade-off is $a=0.5$ and $b=0.1$.

\subsubsection{Different Measures of Statistical Distributions}

To measure knowledge similarity, we use the 2-Wasserstein distance as an approximation of domain similarity (Eqs.~\ref{eq:score} and \ref{eq:fid}). Other common choices for measuring statistical distributions include Maximum Mean Discrepancy (MMD) and Kernel Inception Distance (KID), which are widely used evaluation metrics in generative tasks. However, both MMD and KID require storing features from the original images, whereas the 2-Wasserstein distance only relies on summary statistics (\textit{e.g.}, means and covariances) of the features. Avoiding the need to store or revisit historical images is a key advantage, making the 2-Wasserstein distance more practical for the auto-selection of knowledge adapters. Furthermore, we provide a quantitative comparison of different distance measures. As shown in Tab.~\ref{tab:measure}, the choice of distance metric has only a marginal effect on performance, while the 2-Wasserstein distance offers better practicality.

\re{To further justify the effectiveness and justification of our auto-selection strategy, we compare against non-adaptive selection strategies, including random selection and latest selection. In Tab.~\ref{tab:measure}, both alternatives remove the adaptive selection mechanism, leading to clear performance degradation. This verifies the necessity of adaptive knowledge selection for lifelong person ReID.}

\subsubsection{Layers for Knowledge Adapters}
Inspired by the advanced progress of parameter-efficient fine-tuning~\cite{hu2021lora,fu2023effectiveness,ding2023parameter}, our knowledge adapters can tune the parameters of different layers in the image branch $\mathcal{F}\left(\cdot\right)$. In particular, optional parameterized layers include Query/Key/Value/Projection in the multi-head attention layer as well as the feed-forward network. Therefore, we discuss the necessity to fine-tune them in Tab.~\ref{tab:compare_layer}.
Compared to \#1 and \#2 (or \#3 \textit{v.s.} \#4), we find that removing the knowledge adapter in Key, the generalization of the model can be slightly improved, but it will damage the result on the seen domains. 
By additionally tuning the Projection layer (\#1 \textit{v.s.} \#3 and \#2 \textit{v.s.} \#4), it boosts the performance by about 1 point of mAP and Rank-1 on both seen and unseen domains.
Although introducing FFN provides further gains (\#4 \textit{v.s.} \#6), it doubles storage overhead from 18 MB to 40.5 MB. In this paper, we incrementally add knowledge adapters, like \#4, to layers of the image encoder.

\begin{table*}[t]
  \centering
  \footnotesize
  \caption{\textbf{Comparison with the state-of-the-art methods under LReID setting.} The training order is Market-150 $\rightarrow$ CUHK-SYSU $\rightarrow$ DukeMTMC-reID $\rightarrow$ MSMT17\_V2 $\rightarrow$ CUHK03. 
  ``$\dag$'' denotes rehearsal-based methods using exemplars. ``$\ast$'' indicates reproduced results with released official codes. ``Fine-tuning'' means independently fine-tuning on each dataset. 
  }
   \vspace{-0.05in}
  \label{tab:sota_1}
    \setlength{\tabcolsep}{1.9mm}{
    \begin{tabular}{lccccccccccccccc}
    \toprule
    \multicolumn{1}{c}{\multirow{2}[0]{*}{\textsc{Methods}}} & \multirow{2}[0]{*}{\textsc{Backbone}}
    & \multicolumn{2}{c}{\textbf{MA}} & \multicolumn{2}{c}{\textbf{SY}} & \multicolumn{2}{c}{\textbf{DU}} & \multicolumn{2}{c}{\textbf{MS\_V2}} & \multicolumn{2}{c}{\textbf{CU03}} & \multicolumn{2}{c}{\textbf{\textsc{Seen-Avg.}}} & \multicolumn{2}{c}{\textbf{\textsc{Unseen-Avg.}}}   \\
    \cmidrule(r){3-4} \cmidrule(r){5-6}
    \cmidrule(r){7-8} \cmidrule(r){9-10} \cmidrule(r){11-12} \cmidrule(r){13-14} \cmidrule(r){15-16}
    & & mAP & R-1 & mAP & R-1 & mAP & R-1 & mAP & R-1 & mAP & R-1 & mAP & R-1 & mAP & R-1 \\
    \midrule
    Fine-tuning & ViT-B/16 (CLIP) & 88.8 & 95.3 & 91.6 & 92.9 & 80.4 & 90.0 & 59.7 & 80.6 & 74.1 & 76.7 & 78.9 & 87.1 & - & - \\
    \midrule
    CRL~\cite{zhao2021continual} & ResNet-50 & 58.0 & 78.2 & 72.5 & 75.1 & 28.3 & 45.2 & 6.0 & 15.8 & 37.4 & 39.8 & 40.5 & 50.8 & 44.0 & 41.0  \\
    AKA~\cite{pu2021lifelong} & ResNet-50 & 51.2 & 72.0 & 47.5 & 45.1 & 18.7 & 33.1 & 16.4 & 37.6 & 27.7 & 27.6 & 32.3 & 43.1 & 44.3 & 40.4  \\
    PTKP$^{\dag}$~\cite{ge2022lifelong} & ResNet-50 & 50.3 & 74.8 & 75.4 & 78.0 & 41.2 & 61.5 & 9.8 & 26.3 & 31.7 & 34.1 & 41.7 & 54.9 & 48.8 & 44.5  \\
    PatchKD~\cite{sun2022patch} & ResNet-50 & 68.5 & 85.7 & 75.6 & 78.6 & 33.8 & 50.4 & 6.5 & 17.0 & 34.1 & 36.8 & 43.7 & 53.7 & 49.1 & 45.4  \\
    MEGE$^{\dag}$~\cite{pu2023memorizing} & ResNet-50 & 39.0 & 61.6 & 73.3 & 76.6 & 16.9 & 30.3 & 4.6 & 13.4 & 36.4 & 37.1 & 34.0 & 43.8 & 47.7 & 44.0  \\
    ConRFL$^{\dag}$~\cite{huang2023learning} & ResNet-50 & 59.2 & 78.3 & 82.1 & 84.3 & 45.6 & 61.8 & 12.6 & 30.4 & 51.7 & 53.8 & 50.2 & 61.7 & 57.4 & 52.3 \\
    C$^{2}$R~\cite{cui2024learning} & ResNet-50 & 69.0 & 86.8 & 76.7 & 79.5 & 33.2 & 48.6 & 6.6 & 17.4 & 35.6 & 36.2 & 44.2 & 53.7 & - & - \\
    LSTKC~\cite{xu2024lstkc} & ResNet-50 & 54.7 & 76.0 & 81.1 & 83.4 & 49.4 & 66.2 & 20.0 & 43.2 & 44.7 & 46.5 & 50.0 & 63.1 & 57.0 & 49.9  \\
    LSTKC++~\cite{xu2025long} & ResNet-50 & 65.3 & 83.2 & 85.8 & 87.5 & 56.0 & 70.5 & 21.0 & 43.5 & 47.7 & 48.9 & 55.2 & 66.7 & 63.2 & 56.3  \\
    DKP~\cite{xu2024distribution} & ResNet-50 & 60.3 & 80.6 & 83.6 & 85.4 & 51.6 & 68.4 & 19.7 & 41.8 & 43.6 & 44.2 & 51.8 & 64.1 & 59.2 & 51.6  \\
    DASK~\cite{xu2025dask} & ResNet-50 & 61.2 & 82.3 & 81.9 & 83.7 & 58.5 & 74.6 & 29.1 & 57.6 & 46.2 & 48.1 & 55.4 & 69.3 & 65.3 & 58.4  \\ 
    \midrule  
    PAEMA~\cite{li2024exemplar} & ViT-B/16 & 71.7 & 86.9 & 90.0 & 91.3 & 66.0 & 79.0 & 31.4 & 55.3 & 50.0 & 50.8 & 61.8 & 72.7 & 70.3 & 63.2 \\
    PAEMA$^{\dag}$~\cite{li2024exemplar} & ViT-B/16 & 77.3 & 89.8 & 88.8 & 90.4 & 69.2 & 81.2 & 35.9 & 58.4 & 53.6 & 54.3 & 65.0 & 74.8 & 74.5 & 65.3 \\
    DRE$^{\dag}$~\cite{liu2025diverse} & ViT-B/16 & 69.4 & 85.2 & 85.6 & 86.8 & 52.8 & 67.3 & 21.5 & 43.7 & 54.7 & 58.1 & 56.8 & 68.2 & 56.7 & 55.0 \\
    ATFC~\cite{liu2024attribute} & ViT-B/16 (CLIP) & 75.9 & 87.9 & 87.3 & 88.5 & 60.1 & 71.9 & 25.3 & 50.1 & 60.5 & 61.3 & 61.8 & 71.9 & 60.8 & 58.3 \\
    LSTKC$^{\ast}$~\cite{xu2024lstkc} & ViT-B/16 (CLIP) & 70.2 & 86.4 & 88.7 & 90.3 & 64.2 & 78.3 & 40.9 & 67.4 & \textbf{74.5} & \textbf{76.7} & 67.7 & 79.8 & 74.5 & 67.1 \\
    DKP$^{\ast}$~\cite{xu2024distribution} & ViT-B/16 (CLIP) & 78.0 & 90.8 & 89.7 & 90.3 & 70.1 & 82.3 & 39.6 & 67.0 & 64.5 & 67.4 & 68.3 & 79.2 & 73.5 & 66.2  \\ 
    MoRAL$^{\ast}$~\cite{yang2024moral} & ViT-B/16 (CLIP) & 76.7 & 86.8 & 87.6 & 88.4 & 65.0 & 77.0 & 43.0 & 67.7 & 66.5 & 68.6 & 67.8 & 77.7 & 72.6 & 65.8 \\
    DCR~\cite{liu2025domain} & ViT-B/16 (CLIP) & 75.9 & 87.9 & 87.3 & 88.5 & 60.1 & 71.9 & 25.3 & 50.1 & 60.5 & 61.3 & 61.8 & 71.9 & 60.8 & 58.3 \\
    \midrule
    CLIP-ReID & ViT-B/16 (CLIP) & 64.4 & 82.6 & 84.9 & 86.2 & 62.5 & 78.4 & 39.2 & 65.4 & 71.7 & 74.4 & 64.5 & 77.4 & 69.6 & 62.1 \\
    AdalReID (Ours) & ViT-B/16 (CLIP) & \textbf{86.3} & \textbf{93.7} & \textbf{93.7} & \textbf{94.6} & \textbf{76.0} & \textbf{86.1} & \textbf{56.2} & \textbf{79.2} & 70.3 & 74.2 & \textbf{76.5} & \textbf{85.6} & \textbf{76.8} & \textbf{70.6}  \\
    \bottomrule
    \end{tabular}}
    \vspace{-0.15in}
\end{table*}

\begin{figure}[t]
  \centering
  \includegraphics[width=0.95\linewidth]{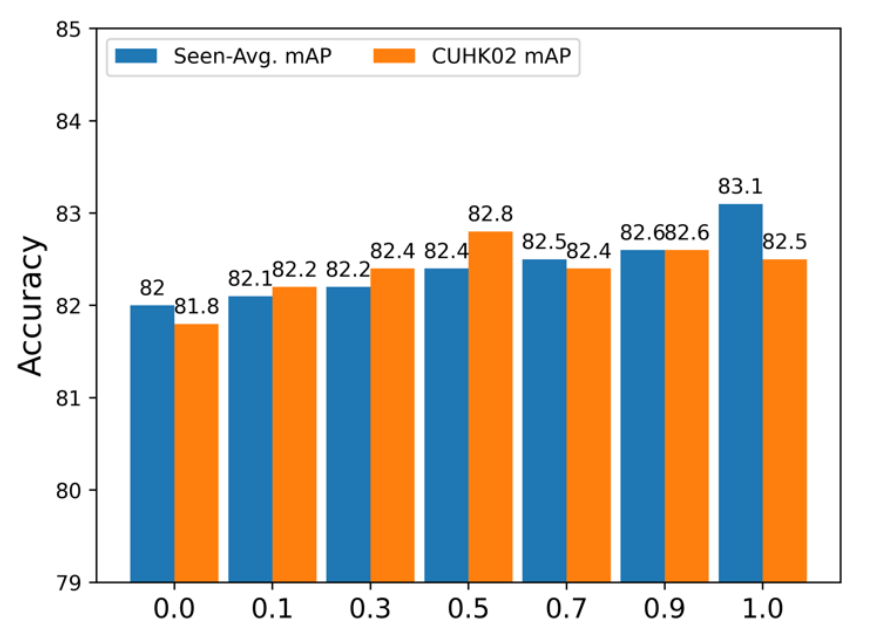}
  \vspace{-0.05in}
  \caption{Ablation studies on mAP performance of different values $\beta$ in knowledge tease-out.
  \label{fig:beta}}
  \vspace{-0.1in}
\end{figure}

\begin{table}[t]
\centering{
\footnotesize{
\setlength{\tabcolsep}{0.3mm}
\caption{
\re{Computational complexity comparisons with existing LReID methods.}
\label{tab:complexity}}
\vspace{-0.05in}
\begin{tabular}{lcccccccc}
    \toprule
     \multicolumn{1}{c}{\multirow{2}[0]{*}{\textbf{\textsc{Metrics}}}} & \multicolumn{4}{c}{\textbf{\textsc{Backbone: ResNet-50}}} & \multicolumn{4}{c}{\textbf{\textsc{Backbone: ViT-B/16 (CLIP)}}}   \\
    \cmidrule(r){2-5} \cmidrule(r){6-9}
     & DKP & LSTKC & ResNet-50 & Ours & DKP & LSTKC & CLIP-ReID & Ours  \\
    \midrule
    Params (M) & 98.4 & 25.1 & 25.1 & 30.9 & 126.8 & 86.1 & 86.1 & 109.7 \\
    FLOPs (G) & 5.6 & 8.1 & 8.1 & 11.5 & 23.5 & 22.8 & 22.8 & 28.8 \\
    \bottomrule
    \end{tabular}}}
    \vspace{-0.05in}
\end{table}

\begin{table}[t]
\centering{
\footnotesize{
\setlength{\tabcolsep}{0.7mm}
\caption{
\re{\textbf{mAP and Rank-1 results using different auto-selection modes.} Order-1 and Order-2 are defined in Tabs.~\ref{tab:seen-orders} and \ref{tab:unseen-orders}. Results are reported after the last training phase.}
\label{tab:sample-wise}}
\vspace{-0.05in}
\begin{tabular}{ccccccccc}
\toprule
\multirow{3}{*}{{{\textsc{Mode}}}} & 
\multicolumn{4}{c}{\makebox[0pt]{\textsc{Order-1}}} & 
\multicolumn{4}{c}{\makebox[0pt]{\textsc{Order-2}}}   \\
\cmidrule(r){2-5} \cmidrule(r){6-9} 
 & \multicolumn{2}{c}{\textbf{\textsc{Seen-AVG.}}} & \multicolumn{2}{c}{\textbf{\textsc{Unseen-AVG.}}} &
 \multicolumn{2}{c}{\textbf{\textsc{Seen-AVG.}}} &
 \multicolumn{2}{c}{\textbf{\textsc{Unseen-AVG.}}} \\
\cmidrule(r){2-3} \cmidrule(r){4-5} \cmidrule(r){6-7} \cmidrule(r){8-9}
& mAP & R-1 & mAP & R-1 & mAP & R-1 &mAP & R-1  \\
\midrule
 Sample-wise & 72.1 & 81.1 & 76.5 & 70.4 & 70.5 & 80.3 & 76.6 & 70.3 \\
Task-wise & 76.5 & 85.6 & 76.8 & 70.6 & 75.3 & 84.4 & 76.8 & 70.5 \\
\bottomrule
\end{tabular}}}
\vspace{-0.2in}
\end{table}

\subsubsection{Efficacy of Knowledge Tease-out}
During training, we merge the weights of different adapters to tease out the required knowledge, a function given by the flexible advantage of our proposed knowledge adapters. Specifically, after the $t$-th stage of training, we update the weight of $W_{o}^{(t-1)}$ via Exponential Moving Average (EMA) in Eq.~\ref{eq:ema}.
To better understand how it works, we experiment with different hyperparameter $\beta$ choices. Several observations can be made in Fig.~\ref{fig:beta}.
\textbf{(i)} On the one hand, we utilize the EMA to sort out knowledge from different tasks, which is a straightforward way of accumulating weighted parameters, \textit{i.e.}, knowledge. As discussed above, it may cause the phenomenon of learning conflict in the shared model. Thus, with the increase of $\beta$, the mAP result in the seen domain is gradually improved.
\textbf{(ii)} On the other hand, EMA is considered to be an effective way to integrate knowledge~\cite{tarvainen2017mean}, especially for historical information. This is exactly the motivation of our proposed knowledge tease-out in Eq.~\ref{eq:ema}. As shown in Fig.~\ref{fig:beta}, the generalization of the model on the unseen domain is improved when the knowledge of historical tasks is introduced into the shared model. The inflection point occurs when $\beta$ is around $0.5$.
\textbf{(iii)} Regardless of the value of $\beta$, the performance is relatively stable in both the seen and unseen domains, which indicates the effectiveness of our proposed auto-selected knowledge adapter and its insensitivity to hyperparameters. Meanwhile, we are expected to select different $\beta$ as needed to achieve further performance gains in practice.

\subsubsection{Analysis of Rank $r$ and Coefficient $\alpha$}
To study the efficacy of knowledge adapters, we conduct experiments by varying different values of rank $r$ and coefficient $\alpha$. These two factors reflect the capacity and importance of our knowledge adapters. Result curves are illustrated in Fig.~\ref{fig:ablation_rank}. We observe that more parameters in knowledge adapters (\textit{i.e.}, from $r=16$ to $64$) will consistently bring about improvements in performance. Besides, appropriately increasing the weight of the adapter can effectively play its role in learning domain-specific knowledge, but when $\alpha$ is large (\textit{e.g.}, $\alpha / r = 8$), there will be a negative impact. Note that we do not consider the selection of $r$ greater than $64$, since it also significantly burdens the storage resources, which is not practical for real-world applications. 
By default, we use $r=64$ and $\alpha=256$ for all experiments.

\begin{table*}[t]
  \centering
  \footnotesize
  \caption{\textbf{Comparison with state-of-the-art methods under LReID setting.} The training order is DukeMTMC-reID $\rightarrow$ MSMT17\_V2 $\rightarrow$ Market-150 $\rightarrow$ CUHK-SYSU $\rightarrow$ CUHK03.
  ``$\dag$'' denotes rehearsal-based methods using exemplars. ``$\ast$'' indicates reproduced results with released official codes. ``Fine-tuning'' means independently fine-tuning on each dataset. 
  }
   \vspace{-0.05in}
  \label{tab:sota_2}
    \setlength{\tabcolsep}{1.9mm}{
    \begin{tabular}{lccccccccccccccc}
    \toprule
    \multicolumn{1}{c}{\multirow{2}[0]{*}{\textsc{Methods}}} & \multirow{2}[0]{*}{\textsc{Backbone}}
    & \multicolumn{2}{c}{\textbf{DU}} & \multicolumn{2}{c}{\textbf{MS\_V2}} & \multicolumn{2}{c}{\textbf{MA}} & \multicolumn{2}{c}{\textbf{SY}} & \multicolumn{2}{c}{\textbf{CU03}} & \multicolumn{2}{c}{\textbf{\textsc{Seen-Avg.}}} & \multicolumn{2}{c}{\textbf{\textsc{Unseen-Avg.}}}   \\
    \cmidrule(r){3-4} \cmidrule(r){5-6}
    \cmidrule(r){7-8} \cmidrule(r){9-10} \cmidrule(r){11-12} \cmidrule(r){13-14} \cmidrule(r){15-16}
    & & mAP & R-1 & mAP & R-1 & mAP & R-1 & mAP & R-1 & mAP & R-1 & mAP & R-1 & mAP & R-1 \\
    \midrule
    Fine-tuning & ViT-B/16 (CLIP) & 80.4 & 90.0 & 59.7 & 80.6 & 88.8 & 95.3 & 91.6 & 92.9 & 74.1 & 76.7 & 78.9 & 87.1 & - & - \\
    \midrule
    CRL~\cite{zhao2021continual} & ResNet-50 & 43.5 & 63.1 & 4.8 & 13.7 & 35.0 & 59.8 & 70.0 & 72.8 & 34.8 & 36.8 & 37.6 & 49.2 & 43.9 & 40.1 \\
    AKA~\cite{pu2021lifelong} & ResNet-50 & 42.2 & 60.1 & 5.4 & 15.1 & 37.2 & 59.8 & 71.2 & 73.9 & 36.9 & 37.9 & 38.6 & 49.4 & 40.6 & 34.2 \\
    PTKP$^{\dag}$~\cite{ge2022lifelong} & ResNet-50 & 34.2 & 52.2 & 7.3 & 19.8 & 55.0 & 78.8 & 79.9 & 81.9 & 30.8 & 41.6 & 41.4 & 54.9 & 48.6 & 43.9  \\
    PatchKD~\cite{sun2022patch} & ResNet-50 & 58.3 & 74.1 & 6.4 & 17.4 & 43.2 & 67.4 & 74.5 & 76.9 & 33.7 & 34.8 & 43.2 & 54.1 & 48.6 & 44.1 \\
    MEGE$^{\dag}$~\cite{pu2023memorizing} & ResNet-50 & 21.6 & 35.5 & 3.0 & 9.3 & 25.0 & 49.8 & 69.9 & 73.1 & 34.7 & 35.1 & 30.8 & 40.6 & 44.3 & 41.1 \\
    ConRFL$^{\dag}$~\cite{huang2023learning} & ResNet-50 & 34.4 & 51.3 & 7.6 & 20.1 & 61.6 & 80.4 & 82.8 & 85.1 & 49.0 & 50.1 & 47.1 & 57.4 & 57.9 & 53.4 \\
    C$^{2}$R~\cite{cui2024learning} & ResNet-50 & 59.7 & 75.0 & 7.3 & 19.2 & 42.4 & 66.5 & 76.0 & 77.8 & 37.8 & 39.3 & 44.7 & 55.6 & - & - \\
    LSTKC~\cite{xu2024lstkc} & ResNet-50 & 49.9 & 67.6 & 14.6 & 34.0 & 55.1 & 76.7 & 82.3 & 83.8 & 46.3 & 48.1 & 49.6 & 62.1 & 57.6 & 49.6 \\
    LSTKC++~\cite{xu2025long} & ResNet-50 & 55.3 & 70.6 & 15.7 & 34.5 & 65.6 & 82.7 & 86.3 & 88.0 & 47.1 & 47.6 & 54.0 & 64.7 & 62.1 & 54.7 \\
    DKP~\cite{xu2024distribution} & ResNet-50 & 53.4 & 70.5 & 14.5 & 33.3 & 60.6 & 81.0 & 83.0 & 84.9 & 45.0 & 46.1 & 51.3 & 63.2 & 59.0 & 51.6  \\ 
    DASK~\cite{xu2025dask} & ResNet-50 & 55.7 & 74.4 & 25.2 & 51.9 & 71.6 & 87.7 & 84.8 & 86.2 & 48.4 & 49.8 & 57.1 & 70.0 & 65.5 & 57.9  \\ 
    \midrule    
    PAEMA~\cite{li2024exemplar} & ViT-B/16 & 67.2 & 79.8 & 26.0 & 49.4 & 69.8 & 85.8 & 89.9 & 91.0 & 49.3 & 49.7 & 60.4 & 71.1 & 69.4 & 62.5 \\
    PAEMA$^{\dag}$~\cite{li2024exemplar} & ViT-B/16 & 68.7 & 80.6 & 29.2 & 52.6 & 72.2 & 57.4 & 90.8 & 91.5 & 52.3 & 51.3 & 62.6 & 66.7 & 71.1 & 63.9 \\ 	 	 
    DRE$^{\dag}$~\cite{liu2025diverse} & ViT-B/16 & 59.7 & 74.2 & 18.7 & 34.8 & 65.4 & 82.7 & 84.8 & 86.7 & 51.9 & 53.2 & 56.1 & 66.3 & 57.1 & 54.9 \\
    ATFC~\cite{liu2024attribute} & ViT-B/16 (CLIP) & 64.1 & 77.2 & 25.4 & 44.9 & 70.6 & 84.5 & 86.1 & 88.2 & 54.2 & 58.7 & 60.1 & 70.7 & 61.6 & 59.2 \\
    LSTKC$^{\ast}$~\cite{xu2024lstkc} & ViT-B/16 (CLIP) & 61.8 & 76.5 & 32.1 & 59.3 & 77.0 & 89.8 & 91.5 & 92.6 & \textbf{75.4} & \textbf{77.3} & 67.6 & 79.1 & 73.8 & 66.1 \\
    DKP$^{\ast}$~\cite{xu2024distribution} & ViT-B/16 (CLIP) & 71.8 & 85.1 & 36.1 & 62.4 & 79.0 & 91.3 & 90.3 & 91.5 & 63.4 & 65.5 & 68.1 & 79.2 & 73.1 & 67.0  \\
    MoRAL$^{\ast}$~\cite{yang2024moral} & ViT-B/16 (CLIP)   
  & 66.7 & 79.2 & 37.0 & 63.2 & 77.9 & 88.0 & 89.2 & 90.2 & 67.3 & 69.9 & 67.6 & 78.1 & 73.4 & 66.9 \\
    DCR~\cite{liu2025domain} & ViT-B/16 (CLIP) & 64.1 & 77.2 & 25.4 & 44.9 & 70.6 & 84.5 & 86.1 & 88.2 & 54.2 & 58.7 & 60.1 & 70.7 & 61.6 & 59.2 \\ 
    \midrule
    CLIP-ReID & ViT-B/16 (CLIP) & 60.1 & 75.6 & 30.1 & 56.4 & 79.1 & 90.7  & 88.9 & 89.9 & 73.6 & 75.3 & 66.4 & 77.6 & 72.6 & 65.9 \\
    AdalReID (Ours) & ViT-B/16 (CLIP) & \textbf{77.9} & \textbf{87.9} & \textbf{51.1} & \textbf{75.2} & \textbf{84.8} & \textbf{93.2} & \textbf{93.2} & \textbf{94.0} & 69.7 & 71.9 & \textbf{75.3} & \textbf{84.4} & \textbf{76.8} & \textbf{70.5}  \\
    \bottomrule
    \end{tabular}}
    \vspace{-0.2in}
\end{table*}

\subsubsection{Limited Testing Samples for Eqs.~\ref{eq:score} and \ref{eq:fid}.}
During testing, our knowledge auto-selector dynamically chooses adapters by quantifying the knowledge similarity between the testing set and the training datasets, as illustrated in Fig.~\ref{fig:ablation_cm}. 
\re{The minimum distance consistently appears along the diagonal, indicating that our proposed similarity measure can reliably distinguish dataset-specific knowledge and identify the most relevant historical adapters for each testing dataset. However, }
this process involves the statistical calculations of the mean and the covariance. A straightforward question is whether the knowledge auto-selector is still reliable if the number of testing samples is limited.
To allay this concern, experimental studies are conducted, and the results in Fig.~\ref{fig:ablation_sample} reveal that the number of images for knowledge similarity indeed affects performance, but it is limited. 
Even with just two testing samples, our model achieves 77.9\% and 82.1\% mAP on seen and unseen domains, compared to 83.1\% and 82.5\% mAP with 100\% of the testing samples. This demonstrates the robustness and practicality of our approach.

\begin{figure}[t]
    \centering
    \includegraphics[width=1.0\linewidth]{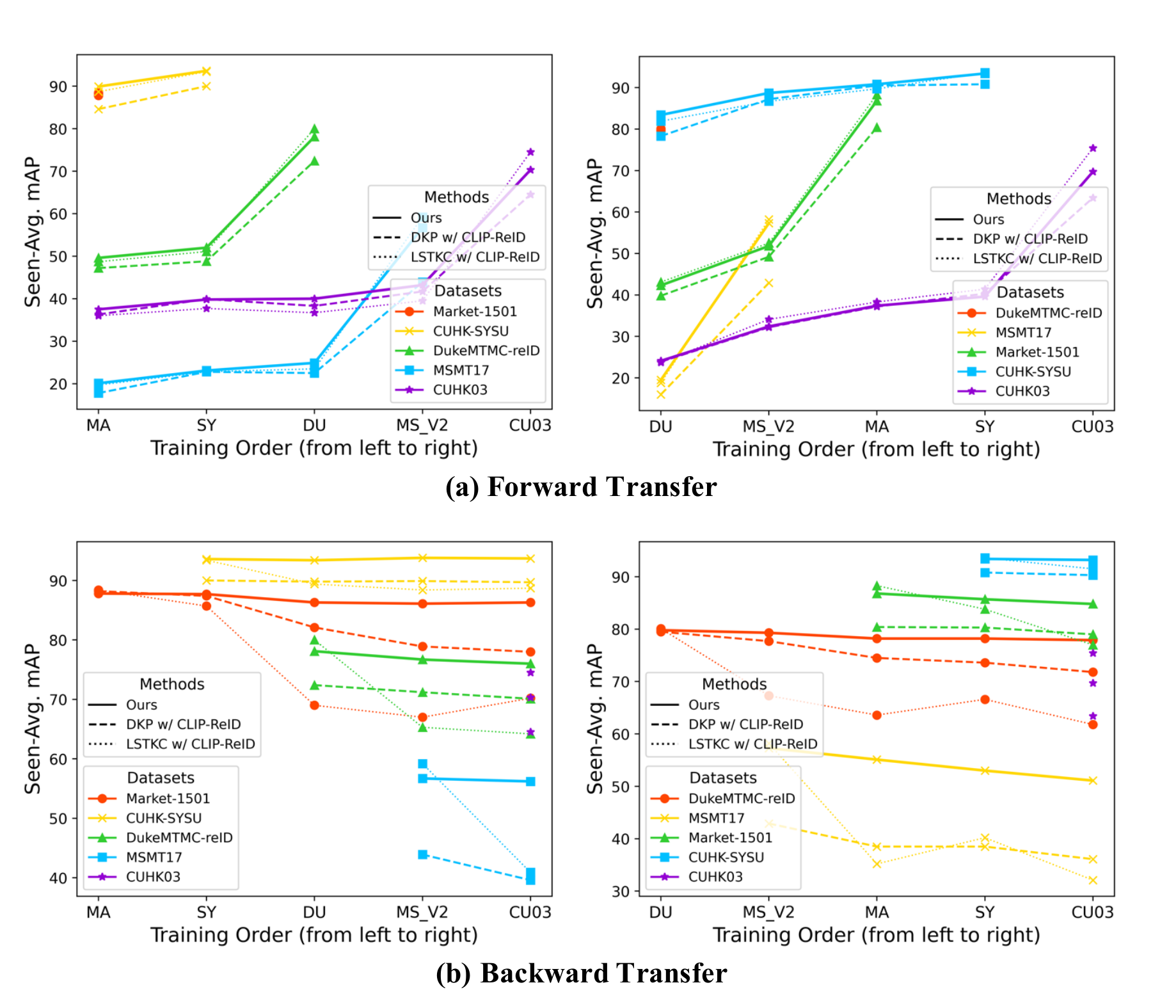}
    \vspace{-0.25in}
    \caption{\textbf{(a) Forward transfer and (b) backward transfer of lifelong learning.} The training sequences for the left and right follow the same orders in Tabs.~\ref{tab:sota_1} and Tabs.~\ref{tab:sota_2}.
    \label{fig:transfer-steps} }
     \vspace{-0.25in}
\end{figure}

\re{For the boundary analysis, we further report the results by choosing adapters in a sample-wise manner. Since only a sample feature vector is available, the covariance term degenerates to a zero matrix. The comparison between task-wise and sample-wise selection is reported in Tab.~\ref{tab:sample-wise}. As expected, using only one sample leads to a less accurate estimation of the feature distribution, resulting in an approximately 4$\sim$5\% mAP drop on the seen domains, while the performance on unseen domains remains nearly unchanged. More importantly, even under this more challenging sample-wise setting, our method still consistently outperforms the competitors reported in Tabs.~\ref{tab:sota_1} and \ref{tab:sota_2}. These results indicate that although task-wise selection provides more reliable adapter selection, the proposed auto-selector remains effective in a sample-wise setting, demonstrating its robustness and practical applicability.}

\subsubsection{Computational Complexity Analysis:} \re{Lastly, we provide a computational complexity analysis comparing our method with existing lifelong person ReID approaches. Since computational complexity (\textit{e.g.}, the number of parameters and FLOPs) requires the source code to measure, we include methods with publicly available implementations, such as DKP~\cite{xu2024distribution}, LSTKC~\cite{xu2024lstkc}, and CLIP-ReID~\cite{li2023clip}. As shown in Tab.~\ref{tab:complexity}, although AdalReID requires additional parameters and computation compared with CLIP-ReID and LSTKC, the introduced overhead is relatively limited considering the benefits for lifelong person ReID. Moreover, under both ResNet-50 and ViT-B/16 backbones, AdalReID possesses fewer parameters than DKP while achieving superior performance on both seen and unseen domains, demonstrating a more favorable balance between model complexity and lifelong learning capability.}

\begin{table*}[t]
\centering{
\footnotesize
\setlength{\tabcolsep}{4.8mm}
\caption{
\textbf{mAP and Rank-1 results using different backbones.} Order-1 and Order-2 are defined in Tabs.~\ref{tab:seen-orders} and \ref{tab:unseen-orders}. Results are reported after the last training phase, and the best is shown in bold.
\label{tab:backbone}}
\vspace{-0.05in}
\begin{tabular}{lccccccccc}
\toprule
\multicolumn{1}{c}{\multirow{3}{*}{{\textsc{Methods}}}} & \multirow{3}{*}{{{\textsc{Backbone}}}} & 
\multicolumn{4}{c}{\makebox[0pt]{\textsc{Order-1}: \textbf{MA $\to$ SY $\to$ DU $\to$ MS\_V2 $\to$ CU03}}} & 
\multicolumn{4}{c}{\makebox[0pt]{\textsc{Order-2}: \textbf{DU $\to$ MS\_V2 $\to$ MA $\to$ SY $\to$ CU03}}}   \\
\cmidrule(r){3-6} \cmidrule(r){7-10} 
 &  & \multicolumn{2}{c}{\textbf{\textsc{Seen-AVG.}}} & \multicolumn{2}{c}{\textbf{\textsc{Unseen-AVG.}}} &
 \multicolumn{2}{c}{\textbf{\textsc{Seen-AVG.}}} &
 \multicolumn{2}{c}{\textbf{\textsc{Unseen-AVG.}}} \\
 \cmidrule(r){3-4} \cmidrule(r){5-6} \cmidrule(r){7-8} \cmidrule(r){9-10}
  &  & mAP & R-1 & mAP & R-1 & mAP & R-1 &mAP & R-1  \\
\midrule
LSTKC~\cite{xu2024lstkc} & ResNet-50  & 50.0 & 63.1 & 57.0 & 49.9 & 49.6 & 62.1 & 57.6 & 49.6 \\
DKP~\cite{xu2024distribution} & ResNet-50 & 51.8 & 64.1 & 59.2 & 51.6 & 51.3 & 63.2 & 59.0 & 51.6 \\
AdalReID (\textit{Ours}) & ResNet-50 & \textbf{62.8} & \textbf{75.1} & \textbf{60.3} & \textbf{53.6} & \textbf{62.1} & \textbf{73.9} & \textbf{60.9} & \textbf{53.1} \\
\midrule
 CLIP-ReID~\cite{li2023clip} & ViT-B/16 (BLIP) & 60.0 & 73.3 & 67.3 & 60.1 & 60.2 & 72.0 & 66.5 & 59.2 \\
 AdalReID (\textit{Ours}) & ViT-B/16 (BLIP) & \textbf{69.7} & \textbf{80.2} & \textbf{69.6} & \textbf{62.3} & \textbf{68.0} & \textbf{79.1} & \textbf{68.4} & \textbf{61.2}\\
\midrule
LSTKC~\cite{xu2024lstkc} & ViT-B/16 (CLIP) & 67.7 & 79.8 & 74.5 & 67.1 & 67.6 & 79.1 & 73.8 & 66.1 \\
DKP~\cite{xu2024distribution} & ViT-B/16 (CLIP) & 68.3 & 79.2 & 73.5 & 66.2 & 68.1 & 79.2 & 73.1 & 67.0 \\
PAEMA~\cite{li2024exemplar} & ViT-B/16 (CLIP) & 71.3 & 81.5 & 75.5 & 69.1 & 69.9 & 80.5 & 75.1 & 68.9 \\
AdalReID (\textit{Ours}) & ViT-B/16 (CLIP) & \textbf{76.5} & \textbf{85.6} & \textbf{76.8} & \textbf{70.6} & \textbf{75.3} & \textbf{84.4} & \textbf{76.8} & \textbf{70.5}\\
\bottomrule
\end{tabular}}
\vspace{-0.15in}
\end{table*}

\begin{table}[t]
  \centering
  \footnotesize
  \caption{\textbf{Comparison with state-of-the-art methods on CIFAR-100 under CIL setting.} The best results are in bold. \label{tab:cil}
  }
   \vspace{-0.2in}
    \setlength{\tabcolsep}{1mm}{
    \begin{tabular}{lcccccccccccccccc}
    \toprule
    \multicolumn{1}{c}{\multirow{2}[0]{*}{\textsc{Methods}}} & \multirow{2}[0]{*}{\textsc{Backbone}}
    & \multicolumn{2}{c}{\textsc{10 Steps}} & \multicolumn{2}{c}{\textsc{20 Steps}} & \multicolumn{2}{c}{\textsc{50 Steps}}   \\
    & & Avg. & Last & Avg. & Last & Avg. & Last  \\
    \midrule
    PRAKA~\cite{shi2023prototype} & ResNet-18 & 70.0 & -- &  68.9 & -- &  65.9 & -- \\
    DyTox+~\cite{Arthur2022dytox} & Transformer & 74.1 & 62.3 & 71.6 & 57.4 & 68.9 & 51.1  \\
    DER~\cite{yan2021dynamically} & ResNet-18 & 74.6 & 64.4 & 74.0 & 62.6 & 72.1 & 59.8\\
    \midrule
    LwF~\cite{li2017learning} & CLIP & 65.9 & 48.0 & 60.6 & 40.6 & 47.7 & 32.9\\
    iCaRL~\cite{rebuffi2017icarl} & CLIP & 79.4 & 71.0 & 73.3 & 64.6 & 71.3 & 59.1  \\
    LwF-VR~\cite{ding2022dont} & CLIP & 78.8 & 70.8 & 74.5 & 63.5 & 71.0 & 59.5  \\
    ZSCL~\cite{zheng2023preventing} & CLIP & 82.2 & 73.7 & 80.4 & 69.6 & 79.9 & 67.4\\
    MoE-Adapters++~\cite{yu2025moe} & CLIP & 85.2 & 77.5 & 83.7 & 76.2 & 83.6 & 75.2\\
    \midrule
    AdalReID (Ours) & CLIP & \textbf{90.5} & \textbf{86.6} & \textbf{86.4} & \textbf{82.2} & \textbf{87.7} & \textbf{80.7}  \\
    \bottomrule
    \end{tabular}}
    \vspace{-0.2in}
\end{table}

\subsection{Comparisons with the State-of-the-Arts \label{subsec:sota}}
Next, we compare AdalReID with LReID methods \cite{pu2023memorizing,xu2024lstkc,li2024exemplar,liu2024attribute,xu2024distribution} and Class-incremental Learning (CIL) approaches \cite{yu2024boosting,zheng2023preventing,rebuffi2017icarl,ding2022dont} to fully investigate its efficacy. 

\subsubsection{Lifelong Person Re-identification Task}
Results of two widely used training orders are represented in Tab.~\ref{tab:sota_1} and \ref{tab:sota_2}. 
Compared with ResNet-based \cite{huang2023learning,cui2024learning,xu2024lstkc,xu2024distribution,xu2025dask} and ViT-based methods \cite{yang2024moral,xu2024distribution,xu2024lstkc,li2024exemplar,liu2025domain,liu2025diverse}, our AdalReID achieves a significant advantage on both the single dataset evaluation and the average results, with approximately 6\% higher on average mAP and Rank-1. A similar superiority of our approach can also be found in the unseen domain evaluation. 
It is worth mentioning that some works \cite{li2024exemplar,xu2024lstkc,yu2023lifelong} set different optimization parameters for the first or last task. We argue that it's a strong prior since defining who is the first or last is ambiguous in practice. By contrast, we keep it the same for all incoming datasets, showing the effectiveness and practicability of our method. 
On the other hand, neither rehearsal-free \cite{xu2024lstkc,xu2024distribution} nor rehearsal-based competitors \cite{huang2023learning,li2024exemplar} yield satisfactory results on MSMT17, which are lower than on other datasets. This is primarily because MSMT17 is notoriously difficult in terms of illumination and scaling, which can lead to discrepancies with ReID knowledge from other datasets, exacerbating the problem of catastrophic forgetting.
Instead, our performance gain on MSMT17 is at least 10\% larger than improvements on other tasks. Thanks to our knowledge adapters and knowledge auto-selectors, we can effectively alleviate this problem while selectively aggregating different knowledge to strengthen generalization. 

\re{Furthermore, we visualize forward and backward transfer results of lifelong learning. 
As shown in Fig.~\ref{fig:transfer-steps}(a), the forward transfer measures whether knowledge acquired from previously learned tasks can facilitate learning subsequent tasks. All three methods exhibit positive forward transfer, indicating that previously acquired knowledge benefits learning of new tasks. Our method achieves forward transfer performance comparable to LSTKC and slightly better than DKP. Figure~\ref{fig:transfer-steps}(b) presents the backward transfer results, which quantify the influence of learning new tasks on the performance of previously learned tasks.
A more stable curve indicates less performance degradation and therefore better resistance to catastrophic forgetting.
As shown, our method exhibits the flattest and most stable curves under both training orders, demonstrating its superior ability to preserve previously acquired knowledge while continually adapting to new tasks. Together, the forward and backward transfer results provide a more comprehensive evaluation of lifelong learning performance and demonstrate that our AdalReID effectively balances knowledge transfer and knowledge retention for lifelong person ReID.}

\begin{table*}[t]
  \centering
  \footnotesize
  \caption{\textbf{Comparison with the state-of-the-art methods.} 
  ``$\dag$'' denotes rehearsal-based methods using exemplars. 
  ``$\ast$'' indicates reproduced results with CLIP backbone. 
  The training order is Market-1501 $\rightarrow$ DukeMTMC-reID $\rightarrow$ CUHK-SYSU $\rightarrow$ MSMT17\_V1, and the results are reported after the last training phase. The best results are shown in bold.}
  \vspace{-0.05in}
  \label{tab:compare_sup1}
    \setlength{\tabcolsep}{3.2mm}{
    \begin{tabular}{lccccccccccccc}
    \toprule
    \multicolumn{1}{c}{\multirow{2}[0]{*}{\textsc{Methods}}}
    & \multicolumn{2}{c}{\textbf{Market-1501}} & \multicolumn{2}{c}{\textbf{DukeMTMC}} & \multicolumn{2}{c}{\textbf{CUHK-SYSU}} & \multicolumn{2}{c}{\textbf{MSMT17\_V1}} & \multicolumn{2}{c}{\textbf{\textsc{Seen-Avg.}}}  & \multicolumn{2}{c}{\textbf{\textsc{Unseen-Avg.}}}   \\
    \cmidrule(r){2-3} \cmidrule(r){4-5} \cmidrule(r){6-7} \cmidrule(r){8-9} \cmidrule(r){10-11} \cmidrule(r){12-13}
    & mAP & R-1 & mAP & R-1 & mAP & R-1 & mAP & R-1 & mAP & R-1 & mAP & R-1 \\
    \midrule
    CRL~\cite{zhao2021continual} & 60.5 & 83.7 & 51.9 & 71.8 & 83.7 & 86.4 & 41.2 & 67.1 & 59.3 & 77.3 & - & 41.6 \\
    GwFReID$^{\dag}$~\cite{wu2021generalising} & 60.9 & 81.6 & 46.7 & 66.5 & 81.4 & 83.9 & 25.9 & 52.4 & 53.7 & 71.1 & - & 40.2 \\
    PTKP$^{\dag}$~\cite{ge2022lifelong} & 77.0 & 90.1 & 63.8 & 78.0 & 88.4 & 90.1 & 41.9 & 67.5 & 67.6 & 81.4 & - & 52.4 \\
    KRC$^{\dag}$~\cite{yu2023lifelong} & 60.3 & 82.3 & 58.7 & 72.7 & 88.9 & 90.5 & 43.3 & 67.7 & 62.8 & 78.3 & - & -\\
    ConRFL$^{\dag}$~\cite{huang2023learning} & 81.4 & 92.8 & 68.4 & 80.5 & 90.0 & 91.5 & 43.0 & 68.5 & 70.7 & 83.3 & 49.2 & 65.7 \\
    Teata~\cite{wang2025image} & 81.0 & 91.8 & 70.7 & 82.1 & 94.0 & 94.8 & 67.4 & 85.7 & 78.2 & 88.6 & - & -\\
    LwF~\cite{li2017learning} + CLIP-ReID~\cite{li2023clip} & 54.0 & 76.2 & 64.7 & 80.7 & 85.6 & 87.5 & 57.7 & 80.1 & 65.5 & 81.1 & - & - \\
    LSTKC$^{\ast}$~\cite{xu2024lstkc} & 63.6 & 83.8 & 64.1 & 78.6 & 89.8 & 91.2 & 71.0 & 87.5 & 72.1 & 85.3 & 65.4 & 69.7  \\
    DKP$^{\ast}$~\cite{xu2024distribution} & 82.2 & 92.5 & 72.7 & 83.8 & 91.7 & 92.7 & 56.1 & 79.4 & 75.7 & 87.1 & \textbf{69.1} & 72.4  \\
    PEAME$^{\ast}$~\cite{li2024exemplar} & 73.1 & 89.5 & 64.9 & 79.4 & 89.9 & 90.9 & 68.4 & 87.1 & 74.1 & 86.7 & 68.4 & 72.2  \\
    \midrule
    CLIP-ReID & 60.9 & 82.0 & 64.2 & 78.9 & 87.6 & 89.3 & 68.0 & \textbf{86.2} & 70.2 & 84.1 & 62.0 & 66.8 \\
    AdalReID (Ours) & \textbf{88.1} & \textbf{94.4} & \textbf{77.9} & \textbf{87.9} & \textbf{95.0} & \textbf{95.8} & \textbf{68.4} & 86.2 & \textbf{82.4} & \textbf{91.1} & \textbf{69.1} & \textbf{72.5} \\
    \bottomrule
    \end{tabular}}
    \vspace{-0.06in}
\end{table*}

\begin{table*}[ht]
  \centering
  \footnotesize
  \caption{\textbf{Comparison with the state-of-the-art methods.} 
  ``$\dag$'' denotes rehearsal-based methods using exemplars. 
  ``$\ast$'' indicates reproduced results with CLIP backbone. 
  The training order is Market-1501 $\rightarrow$ CUHK-SYSU $\rightarrow$ MSMT17\_V2 $\rightarrow$ CUHK03, and the results are reported after the last training phase. The best results are shown in bold.}
  \vspace{-0.05in}
  \label{tab:compare_sup3}
    \setlength{\tabcolsep}{3.7mm}{
    \begin{tabular}{lcccccccccccc}
    \toprule
    \multicolumn{1}{c}{\multirow{2}[0]{*}{\textsc{Methods}}}
    & \multicolumn{2}{c}{\textbf{Market-1501}} & \multicolumn{2}{c}{\textbf{CUHK-SYSU}} & \multicolumn{2}{c}{\textbf{MSMT17\_V2}} & \multicolumn{2}{c}{\textbf{CUHK03}} & \multicolumn{2}{c}{\textbf{\textsc{Seen-Avg.}}} & \multicolumn{2}{c}{\textbf{\textsc{Unseen-Avg.}}}   \\
    \cmidrule(r){2-3} \cmidrule(r){4-5} \cmidrule(r){6-7} \cmidrule(r){8-9} \cmidrule(r){10-11} \cmidrule(r){12-13}
    & mAP & R-1 & mAP & R-1 & mAP & R-1 & mAP & R-1 & mAP & R-1 & mAP & R-1 \\
    \midrule
    LwF~\cite{li2017learning} & 51.7 & 72.8 & 77.9 & 80.8 & 4.9 & 13.6 & 34.1 & 34.1 & 42.2 & 50.3 & 50.6 & 47.5  \\
    iCaRL~\cite{rebuffi2017icarl} & 59.8 & 78.0 & 88.0 & 89.4 & 11.5 & 25.0 & 52.2 & 53.4 & 52.9 & 61.5 & 59.2 & 54.3  \\
    DER++$^{\dag}$~\cite{buzzega2020dark} & 60.4 & 79.2 & 88.9 & 90.5 & 14.2 & 30.0 & 39.5 & 39.5 & 50.8 & 59.8 & 62.6 & 58.1  \\
    AKA~\cite{pu2021lifelong} & 50.7 & 71.3 & 76.8 & 79.8 & 4.8 & 13.5 & 39.5 & 39.5 & 43.0 & 51.0 & 49.4 & 48.4 \\
    MRN~\cite{pu2022metareconciliation} & 67.2 & 85.7 & 84.7 & 86.7 & 18.3 & 41.8 & 42.9 & 43.7 & 53.3 & 64.5 & 60.6 & 56.8  \\
    LSTKC$^{\ast}$~\cite{xu2024lstkc} & 73.5 & 87.9 & 90.8 & 91.7 & 49.9 & 74.1 & 82.6 & 84.8 & 74.2 & 84.6 & 76.3 & 69.7  \\
    DKP$^{\ast}$~\cite{xu2024distribution} & 85.5 & 93.7 & 92.4 & 92.9 & 47.6 & 72.4 & 69.6 & 72.1 & 73.8 & 82.8 & 76.1 & 69.4  \\
    PEAME$^{\ast}$~\cite{li2024exemplar} & 80.8 & 91.6 & 91.3 & 92.2 & 42.6 & 69.1 & 77.6 & 80.0 & 73.1 & 83.2 & 75.8 & 69.5  \\
    \midrule
    CLIP-ReID & 72.8 & 87.1  & 89.7  & 91.2  & 50.5  & 74.8 & \textbf{80.5} & \textbf{83.3} & 73.4 & 84.1 & 74.3 & 67.5 \\
    AdalReID (Ours) & \textbf{89.9} & \textbf{95.1} & \textbf{95.2} & \textbf{95.8} & \textbf{65.7} & \textbf{84.5} & 78.3 & 80.9 & \textbf{82.3} & \textbf{89.1} & \textbf{78.4} & \textbf{72.3} \\
    \bottomrule
    \end{tabular}}
    \vspace{-0.06in}
\end{table*}

\begin{table*}[t]
  \centering
  \footnotesize
  \caption{\textbf{Comparison with the state-of-the-art methods.} 
  ``$\dag$'' denotes rehearsal-based methods using exemplars. 
  ``$\ast$'' indicates reproduced results with CLIP backbone.
  The training order is VIPeR $\rightarrow$ Market-1501 $\rightarrow$ CUHK-SYSU $\rightarrow$ MSMT17\_V2, and the results are reported after the last training phase. The best results are shown in bold.}
  \vspace{-0.05in}
  \label{tab:compare_sup2}
    \setlength{\tabcolsep}{3.7mm}{
    \begin{tabular}{lcccccccccccc}
    \toprule
    \multicolumn{1}{c}{\multirow{2}[0]{*}{\textsc{Methods}}}
    & \multicolumn{2}{c}{\textbf{VIPeR}} & \multicolumn{2}{c}{\textbf{Market-1501}} & \multicolumn{2}{c}{\textbf{CUHK-SYSU}} & \multicolumn{2}{c}{\textbf{MSMT17\_V2}} & \multicolumn{2}{c}{\textbf{\textsc{Seen-Avg.}}}  & \multicolumn{2}{c}{\textbf{\textsc{Unseen-Avg.}}}   \\
    \cmidrule(r){2-3} \cmidrule(r){4-5} \cmidrule(r){6-7} \cmidrule(r){8-9} \cmidrule(r){10-11} \cmidrule(r){12-13}
    & mAP & R-1 & mAP & R-1 & mAP & R-1 & mAP & R-1 & mAP & R-1 & mAP & R-1 \\
    \midrule
    LwF~\cite{li2017learning} & 45.3 & 54.6 & 74.6 & 53.9 & 83.7 & 81.1 & 32.3 & 14.6 & 59.0 & 51.1 & 41.0 & -  \\
    iCaRL~\cite{rebuffi2017icarl} & 56.6 & 67.0 & 78.7 & 56.9 & 82.9 & 80.0 & 24.1 & 10.2 & 60.6 & 53.6 & 42.8 & - \\
    WA~\cite{zhao2020maintaining} & 48.1 & 58.0 & 70.3 & 50.9 & 71.9 & 38.8 & 18.6 & 10.2 & 57.2 & 49.5 & 37.6 & - \\
    AKA~\cite{pu2021lifelong} & 61.7 & 50.6 & 28.3 & 50.7 & 76.9 & 79.6 & 13.4 & 28.0 & 45.1 & 52.2 & 32.6 & -  \\
    PTKP$^{\dag}$~\cite{ge2022lifelong} & 66.3 & 56.0 & 58.3 & 77.4 & 77.1 & 78.8 & 25.2 & 48.0 & 56.7 & 65.1 & 42.2 & - \\
    KRC$^{\dag}$~\cite{yu2023lifelong} & 76.3 & 67.7 & 64.4 & 82.5 & 88.9 & 90.7 & 43.3 & 67.1 & 68.2 & 77.0 & 60.3 & - \\
    DKAR~\cite{chen2026lifelong} & 82.4 & 76.0 & 51.2 & 72.8 & 72.3 & 76.8 & 15.8 & 37.3 & 55.4 & 65.7 & - & 42.1 \\
    LSTKC$^{\ast}$~\cite{xu2024lstkc} & 74.0 & 63.9 & 66.3 & 86.3 & 89.3 & 90.9 & 70.3 & 86.9 & 75.0 & 82.0 & 74.3 & 72.5  \\
    DKP$^{\ast}$~\cite{xu2024distribution} & 83.2 & 75.3 & 79.4 & 91.2 & 92.2 & 92.8 & 57.9 & 80.5 & 78.2 & 84.9 & 76.2 & 74.6  \\
    PEAME$^{\ast}$~\cite{li2024exemplar} & 72.8 & 63.6 & 61.1 & 83.1 & 85.8 & 87.8 & 70.0 & 87.5 & 72.4 & 80.4 & 74.4 & 72.7  \\
    \midrule
    CLIP-ReID & 69.5 & 60.1 & 64.7 & 84.2 & 85.1 & 86.9 & \textbf{68.4} & \textbf{86.5} & 71.9 & 79.4 & 64.5 & 61.9 \\
    AdalReID (Ours) & \textbf{85.6} & \textbf{78.2} & \textbf{84.7} & \textbf{93.1} & \textbf{94.8} & \textbf{95.7} & 66.8 & 85.2 & \textbf{83.0} & \textbf{88.1} & \textbf{76.9} & \textbf{75.1} \\
    \bottomrule
    \end{tabular}}
    \vspace{-0.15in}
\end{table*}

\subsubsection{Class-incremental Learning Task} 
To thoroughly analyze the generalization, we apply our method to the class-incremental learning (CIL) task on the CIFAR100 dataset \cite{krizhevsky2009learning}, dividing the 100 classes into \{10, 20, 50\} subsets and incrementally learning from \{10, 20, 50\} tasks. Following \cite{Arthur2022dytox, yu2024boosting}, we use the average classification accuracy (``Avg.'') and accuracy for the last subset (``Last'') as metrics.
Results in Tab.~\ref{tab:cil} show that our AdalReID is significantly superior to existing CLIP-based methods, even in a much longer 50-step incremental learning regimen. 
In addition, the gap between our ``Avg.'' and ``Last'' accuracies is smaller and more stable across different incremental steps. Note that CIL on CIFAR100 is a classical single-domain lifelong learning task, and our proposed auto-selection strategy still works well, assigning appropriate knowledge adapters for different tasks. These demonstrate the effectiveness and scalability of our proposed method for lifelong learning tasks.

\begin{table*}[t]
  \centering
  \footnotesize
  \caption{\re{\textbf{Comparison with the state-of-the-art methods under LReID setting.} The training order is Market-150 $\rightarrow$ CUHK-SYSU $\rightarrow$ LPW $\rightarrow$ MSMT17\_V2 $\rightarrow$ CUHK03. The best results are shown in bold.}}
   \vspace{-0.05in}
  \label{tab:sota_3}
    \setlength{\tabcolsep}{2mm}{
    \begin{tabular}{lccccccccccccccc}
    \toprule
    \multicolumn{1}{c}{\multirow{2}[0]{*}{\textsc{Methods}}} & \multirow{2}[0]{*}{\textsc{Backbone}}
    & \multicolumn{2}{c}{\textbf{MA}} & \multicolumn{2}{c}{\textbf{SY}} & \multicolumn{2}{c}{\textbf{LPW}} & \multicolumn{2}{c}{\textbf{MS\_V2}} & \multicolumn{2}{c}{\textbf{CU03}} & \multicolumn{2}{c}{\textbf{\textsc{Seen-Avg.}}} & \multicolumn{2}{c}{\textbf{\textsc{Unseen-Avg.}}}   \\
    \cmidrule(r){3-4} \cmidrule(r){5-6}
    \cmidrule(r){7-8} \cmidrule(r){9-10} \cmidrule(r){11-12} \cmidrule(r){13-14} \cmidrule(r){15-16}
    & & mAP & R-1 & mAP & R-1 & mAP & R-1 & mAP & R-1 & mAP & R-1 & mAP & R-1 & mAP & R-1 \\
    \midrule
    LwF~\cite{li2017learning} & ResNet-50 & 44.4 & 68.7 & 66.6 & 69.8 & 25.0 & 36.0 & 6.8 & 19.6 & 52.9 & 53.6 & 39.1 & 49.5 & 50.9 & 43.8\\
    DKP~\cite{xu2024distribution} & ResNet-50 & 60.0 & 80.3 & 84.1 & 85.9 & 46.0 & 57.9 & 17.7 & 38.5 & 41.0 & 41.4 & 49.8 & 60.8 & 57.5 & 50.7  \\
    LSTKC~\cite{xu2024lstkc} & ResNet-50 & 57.0 & 78.6 & 82.9 & 84.9 & 47.2 & 58.4 & 18.4 & 41.1 & 42.3 & 43.7 & 49.6 & 61.3 & 57.8 & 50.2  \\
    DKP++~\cite{xu2024distribution} & ResNet-50 & 52.1 & 76.5 & 75.6 & 77.7 & 45.2 & 58.1 & 14.9 & 36.1 & 40.1 & 42.2 & 45.6 & 58.1 & 56.6 & 48.5 \\
    LSTKC++~\cite{xu2025long} & ResNet-50 & 67.5 & 83.9 & 86.3 & 88.2 & 52.1 & 61.9 & 19.4 & 40.3 & 47.0 & 47.8 & 54.5 & 64.4 & 62.9 & 55.6 \\
    DASK~\cite{xu2025dask} & ResNet-50 & 49.3 & 74.0 & 74.9 & 77.0 & 45.7 & 58.6 & 17.6 & 42.0 & 34.5 & 35.6 & 44.4 & 57.5 & 59.6 & 52.5 \\ 
    \midrule 
    PAEMA~\cite{li2024exemplar} & ViT-B/16 & 73.1 & 88.0 & 90.0 & 91.5 & 59.5 & 68.6 & 30.4 & 54.4 & 48.4 & 48.4 & 60.3 & 70.2 & 69.8 & 63.2  \\
    DRE~\cite{liu2025diverse} & ViT-B/16 & 61.7 & 80.8 & 88.0 & 89.8 & 52.6 & 64.7 & 32.0 & 57.3 & 59.4 & 61.3 & 58.7 & 70.8 & 62.9 & 58.1 \\
    \midrule 
    LwF~\cite{li2017learning} & ViT-B/16 (CLIP) & 51.4 & 77.6 & 73.6 & 76.5 & 36.3 & 46.9 & 13.7 & 35.0 & 27.2 & 26.0 & 40.4 & 52.4 & 55.0 & 47.1 \\
    DKP~\cite{xu2024distribution} & ViT-B/16 (CLIP) & 81.1 & 91.4 & 90.0 & 91.0 & 66.9 & 74.4 & 37.4 & 63.6 & 59.5 & 62.0 & 67.0 & 76.5 & 72.5 & 64.9  \\
    LSTKC~\cite{xu2024lstkc} & ViT-B/16 (CLIP) & 79.8 & 91.4 & 92.0 & 92.7 & 65.8 & 75.1 & 37.8 & 63.5 & 60.6 & 62.4 & 67.2 & 77.0 & 75.3 & 68.5  \\
    PAEMA~\cite{li2024exemplar} & ViT-B/16 (CLIP) & 81.3 & 92.1 & 92.2 & 93.0 & 66.9 & 75.4 & 39.2 & 65.4 & 58.3 & 60.3 & 67.6 & 77.2 & 75.6 & 69.8 \\
    DKP++~\cite{xu2024distribution} & ViT-B/16 (CLIP) & 79.4 & 90.5 & 87.8 & 88.5 & 66.1 & 74.6 & 37.7 & 64.2 & 57.9 & 59.3 & 65.8 & 75.4 & 73.0 & 66.2  \\
    LSTKC++~\cite{xu2025long} & ViT-B/16 (CLIP) & 69.3 & 85.5 & 89.9 & 90.9 & 60.3 & 71.4 & 38.3 & 64.0 & 69.1 & 70.4 & 65.4 & 76.4 & 74.1 & 67.7  \\
    DASK~\cite{xu2025dask} & ViT-B/16 (CLIP) & 79.7 & 90.7 & 91.0 & 91.7 & 64.7 & 73.9 & 42.1 & 67.9 & 63.4 & 65.5 & 68.2 & 78.0 & 74.2 & 66.7  \\ 
    VLADR~\cite{xu2026vision} & ViT-B/16 (CLIP) & 80.1 & 91.9 & 92.0 & 92.5 & 67.8 & 76.2 & 45.5 & 70.5 & 66.6 & 68.4 & 70.4 & 79.9 & 77.8 & \textbf{72.0}  \\ 
    \midrule
    CLIP-ReID & ViT-B/16 (CLIP) & 67.9 & 84.9 & 85.8 & 87.0 & 61.8 & 71.0 & 39.1 & 65.3 & \textbf{74.7} & \textbf{77.0} & 65.9 & 77.1 & 71.6 & 64.6 \\
    AdalReID (Ours) & ViT-B/16 (CLIP) & \textbf{87.2} & \textbf{94.0} & \textbf{93.9} & \textbf{95.0} & \textbf{74.4} & \textbf{82.0} & \textbf{56.5} & \textbf{79.1} & 70.5 & 72.7 & \textbf{76.5} & \textbf{84.6} & \textbf{77.9} & 71.7  \\
    \bottomrule
    \end{tabular}}
    \vspace{-0.06in}
\end{table*}

\begin{table*}[t]
  \centering
  \footnotesize
  \caption{\re{\textbf{Comparison with the state-of-the-art methods under LReID setting.} The training order is LPW $\rightarrow$ MSMT17\_V2 $\rightarrow$ Market-150 $\rightarrow$ CUHK-SYSU $\rightarrow$ CUHK03. The best results are shown in bold.}}
   \vspace{-0.05in}
  \label{tab:sota_4}
    \setlength{\tabcolsep}{2mm}{
    \begin{tabular}{lccccccccccccccc}
    \toprule
    \multicolumn{1}{c}{\multirow{2}[0]{*}{\textsc{Methods}}} & \multirow{2}[0]{*}{\textsc{Backbone}}
    & \multicolumn{2}{c}{\textbf{LPW}} & \multicolumn{2}{c}{\textbf{MS\_V2}} & \multicolumn{2}{c}{\textbf{MA}} & \multicolumn{2}{c}{\textbf{SY}} & \multicolumn{2}{c}{\textbf{CU03}} & \multicolumn{2}{c}{\textbf{\textsc{Seen-Avg.}}} & \multicolumn{2}{c}{\textbf{\textsc{Unseen-Avg.}}}   \\
    \cmidrule(r){3-4} \cmidrule(r){5-6}
    \cmidrule(r){7-8} \cmidrule(r){9-10} \cmidrule(r){11-12} \cmidrule(r){13-14} \cmidrule(r){15-16}
    & & mAP & R-1 & mAP & R-1 & mAP & R-1 & mAP & R-1 & mAP & R-1 & mAP & R-1 & mAP & R-1 \\
    \midrule
    LwF~\cite{li2017learning} & ResNet-50 & 38.8 & 50.5 & 6.7 & 19.2 & 43.9 & 67.1 & 73.2 & 75.9 & 50.1 & 51.9 & 42.5 & 52.9 & 50.0 & 42.7 \\
    DKP~\cite{xu2024distribution} & ResNet-50 & 49.5 & 61.4 & 14.1 & 32.6 & 60.3 & 80.6 & 84.5 & 86.4 & 43.6 & 43.7 & 50.4 & 60.9 & 59.5 & 52.4  \\
    LSTKC~\cite{xu2024lstkc} & ResNet-50 & 46.7 & 57.6 & 14.9 & 33.9 & 56.5 & 78.0 & 84.0 & 86.1 & 42.1 & 43.7 & 48.8 & 59.9 & 57.4 & 49.5  \\
    DKP++~\cite{xu2024distribution} & ResNet-50 & 52.2 & 64.0 & 18.0 & 41.0 & 63.7 & 82.5 & 81.3 & 82.9 & 49.7 & 51.8 & 53.0 & 64.4 & 61.0 & 54.0  \\
    LSTKC++~\cite{xu2025long} & ResNet-50 & 55.7 & 65.7 & 15.0 & 34.3 & 66.9 & 83.6 & 86.3 & 87.6 & 48.4 & 49.9 & 54.4 & 64.2 & 63.4 & 56.0  \\
    DASK~\cite{xu2025dask} & ResNet-50 & 47.2 & 60.4 & 16.1 & 39.9 & 58.7 & 79.9 & 77.5 & 79.4 & 42.0 & 43.0 & 48.3 & 60.5 & 60.0 & 52.8  \\ 
    \midrule 
    PAEMA~\cite{li2024exemplar} & ViT-B/16 & 64.7 & 73.9 & 26.5 & 50.0 & 73.5 & 87.6 & 90.6 & 91.5 & 48.9 & 49.4 & 60.9 & 70.5 & 70.5 & 63.8 \\
    DRE~\cite{liu2025diverse} & ViT-B/16 & 42.2 & 52.3 & 23.6 & 45.8 & 68.2 & 84.3 & 87.3 & 89.1 & 51.9 & 53.2 & 54.6 & 64.9 & 60.0 & 54.4 \\
    \midrule 
    LwF~\cite{li2017learning} & ViT-B/16 (CLIP) & 46.8 & 59.1 & 9.4 & 25.8 & 50.0 & 73.9 & 79.7 & 82.0 & 27.1 & 26.7 & 42.6 & 53.5 & 52.8 & 45.4 \\
    DKP~\cite{xu2024distribution} & ViT-B/16 (CLIP) & 70.2 & 77.5 & 35.7 & 62.1 & 80.0 & 90.7 & 90.9 & 91.5 & 61.9 & 64.0 & 67.7 & 77.2 & 73.9 & 66.7  \\
    LSTKC~\cite{xu2024lstkc} & ViT-B/16 (CLIP) & 69.2 & 77.6 & 34.4 & 60.7 & 74.9 & 88.6 & 92.8 & 93.7 & 60.4 & 61.9 & 66.3 & 76.5 & 74.5 & 68.2  \\
    PAEMA~\cite{li2024exemplar} & ViT-B/16 (CLIP) & 71.2 & 79.2 & 32.4 & 58.8 & 78.1 & 90.5 & 92.9 & 93.5 & 57.7 & 58.2 & 66.5 & 76.0 & 75.3 & 68.8 \\
    DKP++~\cite{xu2024distribution} & ViT-B/16 (CLIP) & 69.1 & 77.5 & 35.3 & 61.9 & 78.6 & 89.9 & 88.5 & 89.2 & 58.9 & 60.5 & 66.1 & 75.8 & 73.2 & 66.6  \\
    LSTKC++~\cite{xu2025long} & ViT-B/16 (CLIP) & 60.7 & 71.0 & 34.5 & 60.6 & 73.3 & 87.0 & 92.1 & 93.0 & 70.6 & 72.9 & 66.2 & 76.9 & 76.0 & 69.7  \\
    DASK~\cite{xu2025dask} & ViT-B/16 (CLIP) & 68.7 & 77.4 & 39.1 & 65.2 & 77.7 & 89.5 & 91.8 & 92.4 & 63.7 & 65.2 & 68.2 & 77.9 & 75.9 & 68.8  \\ 
    VLADR~\cite{xu2026vision} & ViT-B/16 (CLIP) & 69.5 & 77.4 & 40.8 & 67.1 & 80.6 & 91.3 & 92.9 & 93.6 & 67.9 & 69.6 & 70.3 & 79.8 & \textbf{78.1} & \textbf{72.2}  \\ 
    \midrule
    CLIP-ReID & ViT-B/16 (CLIP) & \textbf{80.0} & \textbf{90.9} & \textbf{89.1} & \textbf{90.3} & 60.6 & 69.4 & 31.5 & 58.0 & \textbf{75.7} & \textbf{77.6} & 67.4 & 77.2 & 72.9 & 65.8 \\
    AdalReID (Ours) & ViT-B/16 (CLIP) & 74.8 & 82.0 & 51.8 & 75.6 & \textbf{85.1} & \textbf{93.6} & \textbf{94.1} & \textbf{94.8} & 70.5 & 72.9 & \textbf{75.3} & \textbf{83.8} & 77.4 & 71.3 \\
    \bottomrule
    \end{tabular}}
    \vspace{-0.15in}
\end{table*}

\subsection{Further Scalability Analysis}
Lastly, we explore AdalReID under different training orders and backbones, demonstrating the robustness and scalability of our model across both seen and unseen domains.

\subsubsection{Results with Different Backbones}
In Tabs.~\ref{tab:sota_1} and \ref{tab:sota_2}, we also compare with the baseline (\textit{i.e.}, CLIP-ReID~\cite{li2023clip}) in all experimental settings to show that the efficacy of our approach does not come from the CLIP backbone. 
To further verify this, we conduct experiments to directly evaluate the efficacy of our proposed method on the ResNet-50 backbone. As shown in Tab.~\ref{tab:backbone}, our method consistently achieves higher performance than LSTKC and DKP on both seen and unseen domains, which reveals the efficacy of our proposed method. 
\re{Additionally, our AdaLReID beats the baseline model on the BLIP backbone. Specifically, under Order-1, our method improves the baseline from 60.0\%/73.3\% to 69.7\%/80.2\% (mAP/Rank-1) on the seen domain, while achieving consistent improvements under Order-2. These results demonstrate that our method generalizes well to another representative vision-language backbone.}

\subsubsection{Results with Different Training Orders}
\re{To better verify the efficacy of our proposed AdalReID, we additionally show results with different training orders on Tab.~\ref{tab:compare_sup1}, \ref{tab:compare_sup2}, \ref{tab:compare_sup3}, \ref{tab:sota_3}, and \ref{tab:sota_4}.
Our proposed method consistently outperforms existing approaches across multiple training orders. }
Specifically, AdalReID achieves a clear improvement on the seen domain and competitive performance on the unseen evaluation.
These results demonstrate the robustness and effectiveness of our approach in various evaluation scenarios. The performance gain highlights the superiority of our method.

\section{Limitations and Future Work}
In this paper, we propose a novel framework, AdalReID, that utilizes incremental adapters to learn domain-specific knowledge and introduces the auto-selection strategy to dynamically integrate and assign different adapters for testing datasets. Although it can effectively solve the problem of catastrophic forgetting caused by domain differences, other issues, like LReID across visible and infrared modalities, have not been well studied. 
Directly applying our method to these scenarios inevitably encounters some limitations, which need to be discussed in future work.
On the other hand, a better temperature scheduling dynamic function may lead to more knowledge aggregation and integration, leading to greater generalization capabilities. It would be good future work to explore this direction.

\section{Broader Impact}
Person Re-identification (ReID) technology holds immense potential for transforming various aspects of modern society, significantly impacting public safety, retail, urban planning, and more. 
However, ReID technology must also consider privacy concerns. The ability to continuously track individuals raises ethical questions about surveillance, consent, and data protection. It underscores the need for stringent regulations and transparent practices concerning data collection, storage, and usage to ensure that citizens' rights are respected. 

Lifelong person ReID technology promotes the landing of ReID in its continuous evolution. Multidisciplinary collaborations between technologists, policymakers, and ethicists are still required to maximize societal benefits while mitigating risks. Striking a balance between utility and privacy will be key to realizing the full potential of these technologies in a manner that respects human rights and promotes social good.

It is worth mentioning that all datasets used in our paper are licensed, publicly accessible, and used for research purposes only, except for DukeMTMC-reID~\cite{zheng2017unlabeled}, which has been withdrawn by its creators. However, DukeMTMC-reID is a necessary part of fair person LReID evaluation, since experiments on two major training orders involve this dataset. We use it to facilitate a fair and comprehensive analysis of mainstream benchmarks~\cite{ge2022lifelong,yu2023lifelong,xu2024distribution,xu2024lstkc,liu2024attribute,li2024exemplar,huang2023learning,pu2023memorizing,sun2022patch,xu2025long,xu2025dask}. To protect privacy, we do not identify or display any pedestrian images of DukeMTMC-reID in our paper.
All individuals in the datasets are anonymous with simple identity numbers, and the faces of all person images in our paper are masked.

\section{Conclusion}
In this paper, we propose a novel AdalReID framework with auto-selected knowledge adapters for LReID to mitigate catastrophic forgetting and adapt to evolving identity sets.
AdalReID leverages knowledge adapters to store and integrate old and new knowledge seamlessly.  
Additionally, our parameter-free auto-selection strategy allows for the automatic selection and fusion of knowledge adapters during inference.
Our method achieves new state-of-the-art performance on several benchmarks. 
We anticipate that our method could provide new insights into the scalability of LReID systems and lay a foundation for related research in this field.

\bibliographystyle{IEEEtran}
\bibliography{egbib}

\end{document}